\documentclass{article} % For LaTeX2e
\usepackage{times}

\usepackage[subcaption]{definition}

\usepackage{natbib}
\usepackage{hyperref}
\usepackage{url}
\usepackage{xspace}
\usepackage{graphicx}
\usepackage[utf8]{inputenc} % allow utf-8 input
\usepackage[T1]{fontenc}    % use 8-bit T1 fonts
\usepackage{url}            % simple URL typesetting
\usepackage{booktabs}       % professional-quality tables
\usepackage{tabularx}
\usepackage{multirow}
\usepackage{amsmath,amsfonts,amssymb}
\usepackage{stmaryrd}
\usepackage{nicefrac}       % compact symbols for 1/2, etc.
\usepackage[separate-uncertainty=true]{siunitx}
\usepackage{microtype}      % microtypography
\usepackage[dvipsnames,svgnames]{xcolor}         % colors
\usepackage{float}

\usepackage{svg}

\annotator{ethan}{DarkOrchid}

\newcommand{\ours}{\textsc{CoRGI}\xspace}

\title{\ours \includegraphics[width=0.5cm]{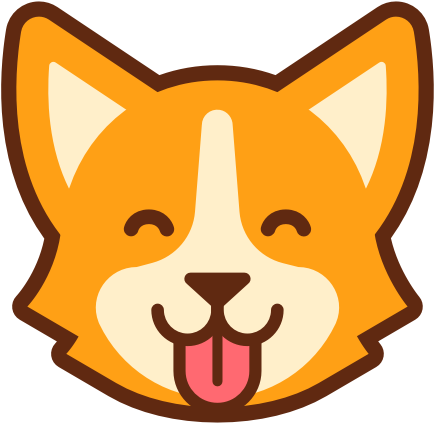}: GNNs with Convolutional Residual Global Interactions for Lagrangian Simulation}

\author{Ethan Ji, Yuanzhou Chen, Arush Ramteke, Fang Sun, Tianrun Yu, Jai Parera, Wei Wang, Yizhou Sun \\
Department of Computer Science\\
UCLA\\
}

\begin{document}

\maketitle

\begin{abstract}
Partial differential equations (PDEs) are central to dynamical systems modeling, particularly in hydrodynamics, where traditional solvers often struggle with nonlinearity and computational cost. Lagrangian neural surrogates such as GNS and SEGNN have emerged as strong alternatives by learning from particle-based simulations. However, these models typically operate with limited receptive fields, making them inaccurate for capturing the inherently global interactions in fluid flows.
% \YS{what is the consequence of missing global and multiscale interactions? Not scalable to large systems? Quality is poor due to missing information?}
Motivated by this observation, we introduce \underline{\textbf{Co}}nvolutional \underline{\textbf{R}}esidual \underline{\textbf{G}}lobal \underline{\textbf{I}}nteractions (\ours), a hybrid architecture that augments any GNN-based solver with a lightweight Eulerian component for global context aggregation. By projecting particle features onto a grid, applying convolutional updates, and mapping them back to the particle domain, \ours captures long-range dependencies without significant overhead.
When applied to a GNS backbone, \ours achieves a 57\% improvement in rollout accuracy with only 13\% more inference time and 31\% more training time. Compared to SEGNN, \ours improves accuracy by 49\% while reducing inference time by 48\% and training time by 30\%. Even under identical runtime constraints, \ours outperforms GNS by 47\% on average, highlighting its versatility and performance on varied compute budgets.
% \YS{emphasize the tradeoff between quality and efficiency can be easily controlled.}
\end{abstract}

\section{Introduction}
% \YS{I like your writing flow of abstract. The best way to arrange introduction is to reuse the same flow but provide more details. The current introduction flow is very dry and way too technical to make people feel interesting.}
\subsection{PDEs and Discretization}
% \YS{the following writing is about both Eulerian and Lagrangian discretization. Please summarize the pros and cons of each method, whichi will lead to the solution of combination of both.}

% Partial differential equations (PDEs) describe the temporal and spatial evolution of dynamical systems, serving as the foundational mathematical models for a wide array of natural and physical phenomena, from hydrodynamics \citep{euler1757principes} and thermodynamics \citep{fourier1822theorie} to electromagnetism \citep{maxwell1873treatise}. Since exact analytical solutions are rarely attainable for real-world systems, numerical solvers have been developed to approximate the behavior of PDEs under discretized space-time representations. These methods typically fall into two paradigms depending on how space is discretized: Eulerian and Lagrangian. 

Partial differential equations (PDEs) provide the governing framework for transport, diffusion, and wave phenomena in continua (e.g. \citet{euler1757principes, maxwell1873treatise, fourier1822theorie}). In fluid mechanics, the conservation of mass, momentum, and energy is expressed by the Navier–Stokes equations. Because closed-form solutions are rarely available for realistic geometries or parameter regimes, computational fluid dynamics (CFD) employs numerical discretizations to approximate these PDEs.

A central modeling choice concerns the kinematic frame, of which two are prevalent. In an \textbf{Eulerian} formulation (e.g. \citep{leveque2002finite, zienkiewicz2005finite, leveque2007finite}), the unknown fields are represented on control volumes fixed in physical space. Conservation laws are enforced through fluxes across the faces of these volumes. In a \textbf{Lagrangian} formulation, the mesh points or particles move with the material velocity. Advection becomes a time derivative along trajectories. For CFD, it is common to use SPH, which computes spatial derivatives through local interpolation among neighboring particles \citep{monaghan1992smoothed, price2012smoothed}. However, these kinematic frames are not mutually exclusive: particle–mesh hybrids such as PIC \citep{evans1957pic}, FLIP \citep{brackbill1986flip}, and MPM \citep{sulsky1994mpm} advect mass or momentum on particles but solve pressure or diffusion on a background grid. As Eulerian methods tend to be weaker at modeling particle-particle interactions and Lagrangian methods tend to struggle on nonlocal constraints, \ours is motivated by combining the strengths of both approaches.

% \textbf{Eulerian methods}, such as finite volume and finite element schemes, discretize the spatial domain using a fixed mesh or grid. These solvers have been the backbone of computational physics and engineering for decades, offering high accuracy and mature numerical stability guarantees \citep{leveque2002finite, zienkiewicz2005finite, leveque2007finite}. However, their reliance on structured grids can make them less effective in scenarios involving moving boundaries, large deformations, or topological changes, requiring complex mesh adaptation or remeshing techniques that introduce additional computational overhead. 

% \textbf{Lagrangian methods}, in contrast, use particles that move with the material flow, discretizing the domain in a mesh-free manner. One of the most widely used Lagrangian approaches is Smoothed Particle Hydrodynamics (SPH), which computes spatial derivatives through local interpolation among neighboring particles \citep{monaghan1992smoothed, price2012smoothed}. While Lagrangian schemes are particularly adept at handling large deformations and topological changes — common in free-surface or multiphase flows — they can be computationally intensive, and may suffer from degraded accuracy in multiscale regimes.

% Eulerian and Lagrangian methods offer complementary strengths and present trade-offs in terms of accuracy, efficiency, and adaptability. This tension motivates the search for novel strategies—particularly in the context of data-driven modeling—that can inherit the advantages of each while mitigating their respective limitations.

\subsection{Machine Learning Surrogates for PDEs}

In recent years, the rise of artificial intelligence in scientific discovery has opened new pathways for solving PDEs using neural PDE surrogates, promising faster and more scalable simulations by learning to approximate PDE solutions. Notable approaches include Physics-Informed Neural Networks (PINNs) \citep{raissi2019physics}, Fourier neural operators (FNOs) \citep{li2020fourier}, and DeepONets \citep{lu2021learning}, which learn mappings between function spaces instead of pointwise solutions. However, most of these efforts target \textbf{Eulerian} representations, where the input structure lends itself naturally to convolutional architectures and structured grids.

Modeling \textbf{Lagrangian} systems requires handling unstructured particle clouds and irregular interactions. Graph neural networks (GNNs) have emerged as a natural fit for this setting, offering the flexibility to learn local interactions between particles \citep{battaglia2016interaction, sanchezgonzalez2020learning}. LagrangeBench \citep{toshev2023lagrangebench} introduced a benchmark suite for evaluating GNN-based surrogates on various fluid systems, including dam breaks, lid-driven cavities, and Taylor-Green vortices, providing a much-needed standardization in this space. Despite their expressiveness, GNNs are computationally limited: long-range interactions require deep GNN stacks, which are prone to over-smoothing and memory inefficiency.
% (see Section??? for a technical discussion). 

\subsection{\ours: A Hybrid Architecture}

Traditional solvers have long recognized the strengths of combining Eulerian and Lagrangian discretizations. While Lagrangian methods excel at local accuracy and topological flexibility, Eulerian grids offer efficient global operations and regular data structures. 
% However, to the best of our knowledge, this synthesis remains underexplored in the context of neural PDE surrogates.
In this work, we pursue a neural hybrid approach: \textbf{an Eulerian augmentation of Lagrangian GNNs}. Our key observation is that local feature aggregation — well handled by GNNs — scales well in large systems due to the spatial locality of interactions, while the bottleneck of global aggregation can be addressed through grid-based projection of particle features. In particular, our intuition is that emulation of SPH may fail to capture global dynamics (e.g. inverse energy cascades) and that emulation of MPM may fail to capture local dynamics (i.e. particle-particle interactions), but our method synthesizes both techniques without being computationally prohibitive. To our knowledge, such a synthesis remains underexplored in the traditional methods literature, and as a result, is similarly untapped within the neural surrogates literature.

% Of particular significance is the global interactions within hydrodynamics (Sec~\ref{sec:cfd}). In many other PDE systems, interactions remain largely local: effects of distant entities are lost by diffusion into noise. However, in hydrodynamics (e.g. meteorology, oceanology, geophysics, etc.), the assumption of locality is challenged by advection. Although the systems may still be considered local, in the sense that molecular electromagnetic repulsion is a local phenomenon, in practice it is useful to model these interactions non-locally; we discuss the rationale in more depth in Sec~\ref{sec:disc}.

We introduce \underline{\textbf{Co}}nvolutional \underline{\textbf{R}}esidual \underline{\textbf{G}}lobal \underline{\textbf{I}}nteractions (\ours), a CNN-driven graph neural network variant that enhances their ability to model long-range interactions in Lagrangian PDE systems. \ours projects intermediate graph features onto an Eulerian grid, performs efficient global aggregation using convolutions, and maps the result back to the particle domain for downstream processing. This design retains the local modeling power of GNNs while offloading global feature handling to CNNs, resulting in a hybrid surrogate that is both expressive and efficient.

Our main contributions are as follows:
\begin{itemize}
    \item We identify a computational bottleneck in GNN-based Lagrangian surrogates arising from long-range interactions, and addresses it by proposing a novel hybrid architecture \ours that augments GNNs with CNN-based global aggregation via a learned Eulerian transformation. 
    \item We validate \ours through extensive experiments on the LagrangeBench suite, demonstrating that \ours on average doubles rollout prediction accuracy across all benchmark tasks, with 13\% runtime overhead.
    \item We further demonstrate \ours enables shallower, faster GNNs without performance decrease: \ours performs on average 47\% better than GNS with the same time budget.
    % \item We offer a conceptual bridge between traditional hybrid discretizations and their neural counterparts, setting a direction for future research in neural scientific computing. 
\end{itemize}

\subsection{Related Works}
For our model architecture, we conducted a survey of the literature to determine both our Lagrangian and our Eulerian components. 
% \wei{because GNS and SEGNN are the two baseline models we will furnish comparison, we need to present what they are, their key ideas, major limitations, difference to our approach... in the related work}\ethan{I introduce them briefly in lagrangian section, not sure if it is sufficient}

\textbf{Eulerian surrogates.} A U-Net \citep{ronneberger2015unet} based Eulerian surrogate, compared to many related methods (e.g. FNO \citep{li2021fourier}, F-FNO \citep{tran2021factorized}, GINO \citep{li2023geometry}, transformer-based neural operators \citep{cao2021choose, hao2023gnot, li2022transformer, hao2024dpot, wang2024bridging}), was found to be more efficient for a given input resolution during inference time by \citet{alkin2025universal}. Other Eulerian methods present other challenges: 
% UPT \citep{alkin2025universal} would dominate training time (the authors claim training UPT-8M takes 150 A100 hours; \ours 
% % \wei{we need to properly present GNS, what is the key idea in it, major limitations, before we use it} 
% takes 5 to 20 A100 hours, depending on dataset and hyperparameters); 
MeshGraphNets \citep{pfaff2021learning} are known to struggle with non-local features; its extension to multiscale \citep{fortunato2022multiscale} requires the generation of additional meshes. We utilize a grid for this coarser mesh, which is more amenable to efficient convolution operations than other types of meshes. We use a grid-based convolution instead of a continuous convolution \citep{ummenhofer2019lagrangian} as they were found to perform better at a lower computational cost by \citet{rochman-sharabi2025a}. Nonetheless, some Eulerian or hybrid methods are comparable to ours, and hence we provide comparisons to UPT \citep{alkin2025universal} and NeuralMPM \citep{rochman-sharabi2025a} in App~\ref{app:abl}.

As discussed in App~\ref{sec:lim}, using efficient adaptive mesh refinement while preserving high efficiency remains an open challenge. We leave the integration of the more costly surrogates into the \ours framework for future work.

\textbf{Lagrangian surrogates.} We identify two architectures considered state of the art for Lagrangian simulations: GNS \citep{sanchezgonzalez2020learning} and SEGNN \citep{brandstetter2022geometric}. GNS is a neural simulator which learns graph message passing in order to predict dynamics from a graph of particles; we elaborate upon this in Sec~\ref{sec:gns}. SEGNN is another simulator which enforces equivariance in its update and message passing functions and utilizes steerable feature vectors for representing covariant information. GNS \citep{sanchezgonzalez2020learning} is considered the fastest Lagrangian surrogate due to its simplicity \citep{toshev2023lagrangebench}, and hence serves as our backbone in the main text. Many popular architectures \citep{brandstetter2022geometric, thomas2018tensor, anderson2019cormorant, Batzner_2022} rely on expensive Clebsch-Gordan tensor products, but we nonetheless also provide comparisons (Table~\ref{tab:results}) to SEGNN \citep{brandstetter2022geometric}, demonstrating its inefficiency compared to \ours.

We observe that these models only facilitate local interactions. We note that a failure mode of these models is their poor enforcement of fluid incompressibility (Fig~\ref{fig:dam}). We hypothesize this is due to the low propagation speed of purely local GNNs, which we address in \ours by increasing the receptive field of each node, thus increasing the propagation speed of our architecture without increasing temporal resolution. Given the improved accuracy using \ours compared to the other architectures, we identify \ours as Pareto optimal.

To our knowledge, the most similar contribution in the literature to \ours is Neural SPH \citep{toshev2024neural}, which introduces spherical particle hydrodynamics relaxation steps during inference time. However, Neural SPH does not change the neural model itself, only providing an inference time adjustment. In addition, this adjustment relies on normalization statistics, making it brittle in unknown scenarios. Neural SPH and \ours are orthogonal augmentations to Lagrangian neural networks: Neural SPH applies a local correction, while \ours aims to capture global features. As we improve different aspects of neural surrogates, these approaches are complementary, i.e. it is possible to apply both \ours and Neural SPH to GNNs.

\section{Preliminaries}
\label{sec:pre}

\textbf{Notations.} In the text, we utilize Hadamard notation frequently in representing our operations, similar to what one might expect in various computer programming contexts. 
We generally use regular lower case letters to denote scalars, bold lower case letters to denote vectors, and bold upper case letters to denote matrices and high-dimensional tensors. 
% We also differentiate between values pertinent to the whole graph (e.g. a variable representing every node) by bolding it, whereas values representing only a particle are left unbolded. 
We represent the concatenation of features $\mathbf{x}$ and $\mathbf{y}$ as $\mathbf{x} \oplus \mathbf{y}$, and the composition of $f$ and $g$ as $f \circ g$. For integers $a < b$, we use $\llbracket a, b \rrbracket$ to denote the set of integers $\{ a, a+1, \cdots, b \}$. Our variables include $i, j \in \llbracket 1, N\rrbracket$ to enumerate particles, $\ell$ to enumerate message passing steps, and $k$ to represent the number of levels for convolution. All other variables are introduced as they appear in the text. 

\subsection{Computational Fluid Dynamics (CFD)}
\label{sec:cfd}
We specifically focus on applying our method to CFD simulations, which are governed by the Navier-Stokes equation. Unlike the heat or wave equation, the Navier-Stokes equation is a nonlinear PDE and is more challenging to simulate. The general form of the equation is
\[\rho \left(\frac{\partial}{\partial t} + \mathbf{u} \cdot \nabla\right)\mathbf{u} = -\nabla p + \nabla \cdot \boldsymbol{\tau} + \rho \mathbf{a}\] where $\rho$ is the density, $\mathbf{u}$ is the flow velocity, $p$ is the pressure, $t$ is time, $\boldsymbol{\tau}$ is the Cauchy stress tensor, and $\mathbf{a}$ represents externally induced acceleration, e.g. gravity. In the incompressible regime, $\boldsymbol{\tau} = \mu\left[\nabla\mathbf{u} + (\nabla\mathbf{u})^{\top}\right]$, where $\mu$ is the dynamic viscosity; we note that incompressibility implies being divergence free, i.e. $\nabla\cdot\mathbf{u}=0$. The quadratic advective term $(\mathbf{u} \cdot \nabla) \mathbf{u}$ invites both local and global scale phenomena; as it is nonlinear, this prevents solutions for different parts of the flow from being independent of one another: for instance, in 2D flows, it is common to observe inverse energy cascades, i.e. energy is transferred from smaller scales to larger scales. We wish to incorporate this mathematical prior by utilizing a neural surrogate for both local and global modes; methods like SPH by themselves struggle with larger scales, as the diffusion and pressure projection steps require large neighborhoods to be accurate.

% As a general solution to the Navier-Stokes equation is unknown, it is not possible to compute trajectories analytically. By replacing the continuum with a finite set of degrees of freedom, CFD renders otherwise intractable flows amenable to simulation and analysis. Two complementary discretization philosophies dominate the field:

% \textbf{Lagrangian Discretization.}\quad
% A Lagrangian method tracks individual material parcels as they convect, thereby eliminating the explicit advection term and naturally enforcing mass conservation. There are 3 main families of Lagrangian techniques: discrete element methods \citep{cundall1979discrete}, material point methods \citep{brackbill1986flip}, and smoothed particle hydrodynamics \citep{gingold1977smoothed}.

% \textbf{Eulerian Discretization.}\quad An Eulerian approach fixes control volumes in space and solves balance laws for the fluid that moves through them. Concretely, instead of tracking attributes for each particle as in Lagrangian discretization, Eulerian methods track attributes for fixed grid or mesh points.

\subsection{Problem Statement}

Our aim is to autoregressively estimate future positions of particles $\hat{\mathbf{x}} \in \mathbb{R}^{N \times d}$. To do so, we define a graph network surrogate that predicts their acceleration, i.e. $\mathcal{F}_\theta : (\mathcal{G}_0, \mathbf{x}_0, \mathbf{p}) \mapsto \hat{\ddot{\mathbf{x}}} \in \mathbb{R}^{N \times d}$, where $N \in \mathbb{N}$ is the number of particles, $d \in \{2, 3\}$ is the spatial dimension, $\mathbf{x}_0 := (x_i : i \in \llbracket 1, N\rrbracket) \in \mathbb{R}^{N \times d}$ is the current particle positions, $\mathcal{G}_0 = (\mathcal{V}, \mathcal{E})$ is a graph (such that $\mathcal{V}$ represents particles and $\mathcal{E}$ is constructed based on distance) where $|\mathcal{V}| = N$, $\mathbf{p} \in \llbracket 1, T\rrbracket^N$ is the particle types, and $\hat{\ddot{\mathbf{x}}}$ is the predicted acceleration. Then, positions at the following time-step are estimated by taking the symplectic Euler integral \citep{vogelaere1900methods}: $
 \hat{\dot{\mathbf{x}}}^{(t+1)} = \hat{\dot{\mathbf{x}}}^{(t)} + \Delta t \hat{\ddot{\mathbf{x}}}^{(t)};\space \hat{\mathbf{x}}^{(t+1)} = \hat{\mathbf{x}}^{(t)} + \Delta t \hat{\dot{\mathbf{x}}}^{(t+1)}. $

\subsection{Graph Encoding} \label{sec:gcon}
The multilayer perceptrons (MLP) \citep{rumelhart1986learning} $\varphi$ used below are the standard implementation with ReLU activation \citep{agarap2019deep} and LayerNorm normalization \citep{ba2016layer}, hence their mathematical formulations are not provided. Similarly, we use the standard convolution \citep{lecun1998gradient} and transposed convolution \citep{dumoulin2016guide} layers. We use a stride of $2$ for pooling operations unless stated otherwise.
% \YS{the previous paragraph is not about graph encoding. Move to somewhere else?}\ethan{I can move to implementation details section in appendix}

We use a graph to facilitate particle-particle/local interactions.
% \YS{add one sentence here to indicate why graph encoding is needed, and connection to other components.} 
We consider the graph at time $t$ denoted by $\mathcal{G}=(\mathcal{V},\mathcal{E}, \{\bh_i^0\} _{i\in\mathcal{V}}, \{\be_{ij}^0\} _{(i,j)\in\mathcal{E}})$, where 
i.e. an object with $\mathcal{V}$ denotes vertices, $\mathcal{E}$ denotes edges, and $\bh_i^0$, $\be_{ij}^0$ denote node and edge features, respectively. 
% \YS{Is your graph dynamic and based on the position of each node? If so, shall we consider using notation $G^t$? } 
We use initial node and edge features 
\begin{equation}
\bh_i^{0}= \Bigl(
  \xb_{i}^{t-\mathcal{L}:t} \oplus
  \tilde{\dot{\xb}}_{i}^{t-\mathcal{L}:t-1} \oplus
  \lVert \dot{\xb}_{i}^{t-\mathcal{L}:t-1}\rVert \oplus
  b_i \oplus
  \bbf_i \oplus
  \bt_{p_i}
\Bigr) \in\mathbb{R}^{F_n}
\end{equation}
\begin{equation}
\be_{ij}^{0}=\Bigl(
    (\xb_i^{t}-\xb_j^{t}) \oplus
    \lVert \xb_i^{t}-\xb_j^{t}\rVert
\Bigr)\in\mathbb{R}^{F_e}
\end{equation}
where $\mathcal{L} \in \mathbb{N}$ is the length of history we pass to the model, $\xb_{i}^{t-\mathcal{L}:t} \in \mathbb{R}^{d \times \mathcal{L}}$ is the position history, $\tilde{\dot{\xb}}_{i}^{t-\mathcal{L}:t-1} \in \mathbb{R}^{d \times (\mathcal{L} - 1)}$ is the direction history (expressed as unit vectors), $\lVert \dot{\xb}_i^{t-\mathcal{L}:t-1}\rVert \in \mathbb{R}^{d \times (\mathcal{L} - 1)}$ is the speed history, $b_i \in \mathbb{R}$ is the distance from a boundary, $\bbf_i \in \mathbb{R}^d$ is the external forces vector, and $\bt_{p_i} \in \mathbb{R}^{F_t}$ is the learned type embedding. For datasets with only one particle type, we omit $\bt_{p_i}$ from the node features. 
% \YS{$\oplus$ denotes concatenation? }\ethan{yes}\YS{add it to the text please.}\ethan{it is already in the notation section}

To complete the encoding, two MLPs, $\varphi_n:\mathbb{R}^{F_n}\to\mathbb{R}^H$ and $\varphi_e:\mathbb{R}^{F_e}\to\mathbb{R}^H$, yield latent embeddings $\bh_i=\varphi_n( \bh_i^0)$, $\be_{ij}= \varphi_e(\be_{ij}^0)$.

\subsection{Graph Network-based Simulators (GNS)} \label{sec:gns}
The GNS is a graph-based Lagrangian surrogate which learns message passing for predicting dynamics introduced by \citet{sanchezgonzalez2020learning}. 
% \YS{add a sentence to cite and introduce GNS before going to implementation.} 
For each layer of the GNS, the latent graph is updated by 2 MLP layers with residual connections. Formally, for $\ell \in \llbracket 1, L \rrbracket$, where $L \in \mathbb{N}$ is the number of message passing steps, $\mathcal{G}^{(\ell)} = \Psi^{(\ell)}(\mathcal{G}^{(\ell - 1)})$,
\begin{align}
\Psi^{(\ell)} := \left(1 + \varphi_n^{(\ell)} \circ \mathcal{U}_n\right)\circ\left(1 + \varphi_e^{(\ell)}\circ\mathcal{U}_e\right)
\end{align}
where $\varphi_e^{(\ell)} : \mathbb{R}^{3H} \rightarrow \mathbb{R}^H$ and $\varphi_n^{(\ell)} : \mathbb{R}^{2H} \rightarrow \mathbb{R}^H$ are both MLPs, $\mathcal{U}_e$ is the edge update map $\left\{\bh_i, \bh_j, \be_{ij}\right\} \mapsto \bh_i \oplus \bh_j \oplus \be_{ij}$, and $\mathcal{U}_n$ is the node update map $\{\bh_i\} \cup \left\{\be_{ij}\right\}_{j:(i,j)\in\mathcal{E}} \mapsto \bh_i \oplus \sum_{j:(i,j)\in\mathcal{E}} \be_{ij}$. This is the implementation found in \citet{toshev2023lagrangebench}, and hence the implementation used for our baseline and our augmentation. We provide the mathematical formulation here, as it slightly differs from the original formulation from \citet{sanchezgonzalez2020learning}. However, GNS models only the local interactions explicitly.
% \YS{add a sentence here introducing the limitation, and transition to the next section on our own method.}
\section{Methodology} \label{sec:methodology}

\begin{figure}[t]
    \centering
    \includegraphics[width=\linewidth]{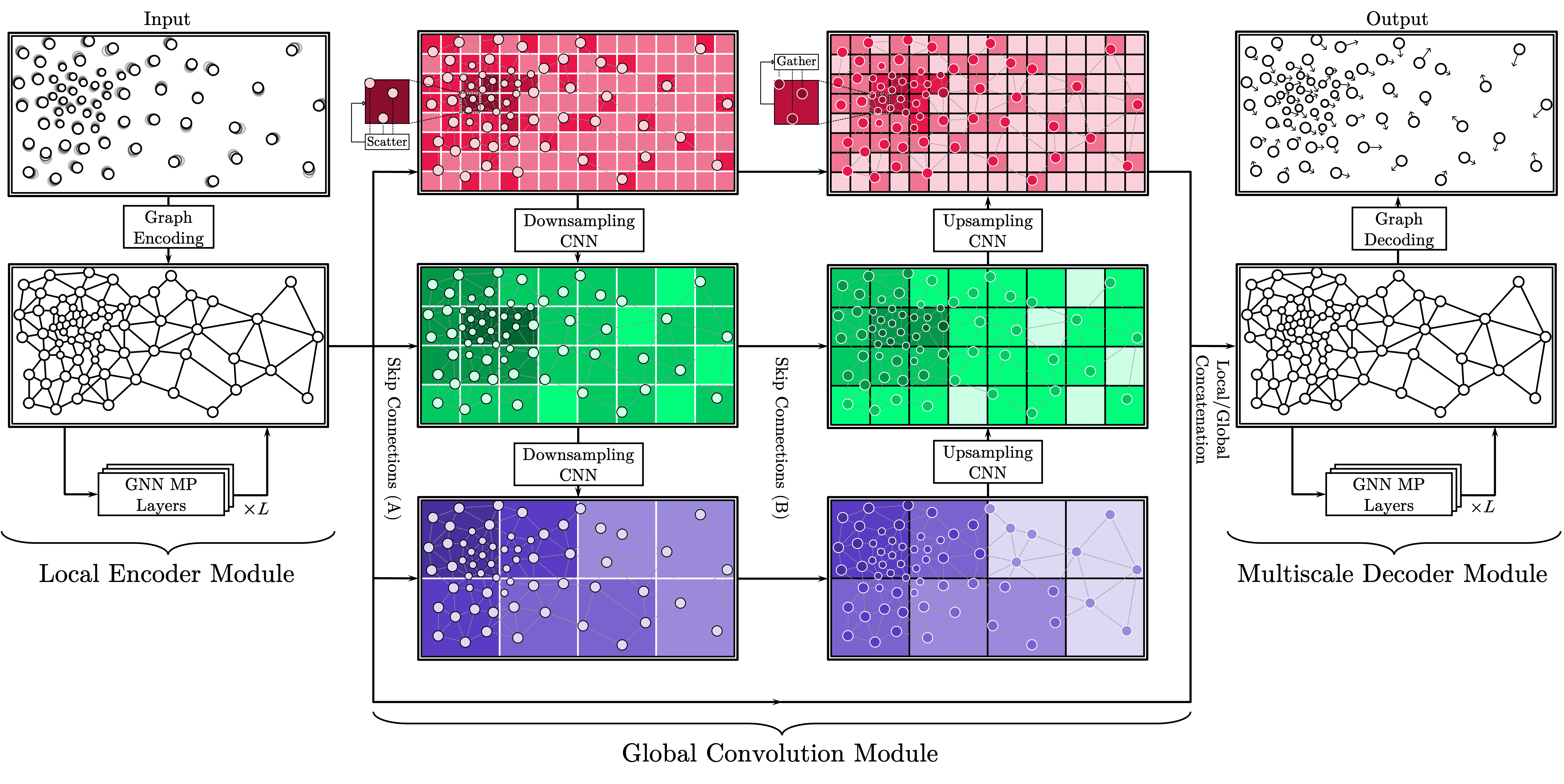}
    \caption{An illustration of the \ours architecture over an example 2D particle dataset. 
    For a GNN with $2L$ layers, our method takes node features encoded by the GNN at layer $L$, processes them into grid-based features at different resolution levels through a scatter operation, and then apply a multi-resolution CNN inspired by U-net \citep{ronneberger2015unet} to capture global features. After this, the processed grid feature at the highest resolution are mapped back to node features, and finally the latter $L$ layers of GNN message passing decodes the features into outputs. 
    % \YS{are scatter and gather the right words here? To me it is more gather and scatter, or pooling and unpooling} \ethan{They are the right words}
    % \arush{An overall walkthrough in this caption would be nice, or refer to section 3 if added there}
    } \label{fig:arch}
\end{figure}

%\wei{add a couple of sentences to describe the overall process, highlight the key contribution we bring.} 
% \YS{transition from previous section from GNS} 
We observe that hydrodynamics often exhibits multiscale phenomena, and hence we wish to model nonlocal interactions. In particular, advection facilitates acoustically propagated local interactions, which are often approximated as nonlocal interactions as discussed in Sec~\ref{sec:disc}. Unfortunately, increasing the interaction range scales computational costs by $\mathcal{O}(r^d)$. 
% \YS{add the consequence if we increase the interaction range at the GNN framework (from both efficiency and quality perspective)}
Hence, the contribution of \ours is to incorporate global computation and information into Lagrangian surrogates via integrating with the Eulerian representation. We first generate representations utilizing a GNN encoder, then we perform convolution to expand a particle's receptive field, and lastly we fuse both the local and global information and utilize a GNN decoder. We describe in more detail the pipeline of \ours in Fig~\ref{fig:arch}. 
% \YS{the figure is too far away. Do you want to create another smaller figure to replace fig 1 to illustrate the two types of discretization? }

\subsection{Full Architecture of \ours}
% \YS{3.1 can be enriched together with Fig. 1, with notations and high-level formulas.}

\textbf{GNN-based Local Encoder for Particles.} We utilize the architecture outlined in Sec~\ref{sec:gns} as an encoder with $L$ layers. Intuitively, it is desirable to have some topological and local geometric priors before we perform convolution. We denote its output $\bh^\downarrow_i$.

\textbf{Grid Encoding via Scattering.} To assign values to our lattice, we use cloud in cell scattering \citep{birdsall1969clouds} on $\bh^\downarrow_i$ to create a grid (referred to later as $G_0^{(1)}$). A precise formulation of this may be found in App~\ref{app:intpl}. 
% \YS{notations are needed to make it precise. What to represent and with which embedding?}\ethan{NeurIPS reviewers said they thought it was too confusing with the notation… I moved to appendix so they can ignore it if they want, but I do agree that it is good to have}

\textbf{Convolutional Global Interaction.}
We use a convolutional component inspired by U-Net \citep{ronneberger2015unet} to allow for global feature interactions across long distances.

Given a tuple of CNN widths $(F_1, \dots, F_K)$, we apply a convolution block $K$ times for downsampling: 
\begin{equation}
G^{(k)} = \text{ConvBlock}_{F_k}\left(G^{(k-1)} \oplus G_0^{(k-1)}\right)
\end{equation}
where $G^{(k)}_0 = \text{AvgPool}\left(G_0^{(k-1)}\right)$ and $G^{(1)} = G_0^{(1)}$ is the output from the last step. For upsampling, we take $\tilde{G}^{(k-1)} = \text{ConvBlock}_{F_k}\left(\text{UpConv}\left(\tilde{G}^{(k)} \oplus G^{(k-1)}\right)\right)$. The output is $\tilde{G} = \tilde{G}^{(1)} \in \mathbb{R}^{|\mathcal{C}|\times H}$. The convolution blocks above are standard for U-Net: double convolution using ReLU activation and InstanceNorm normalization \citep{ulyanov2017instance}.

\textbf{Conversion to Particles via Gathering.} We use cloud in cell gathering on $\tilde{G}$, yielding particle vectors $\tilde{\bh}_i$. We then augment the particle representations $\bh_i^\text{aug} = \bh^\downarrow_i \oplus \tilde{\bh}_i$.
% \YS{this part needs more explanation. input and output? Can anyone implement the algorithm by reading this paragraph? Probably not. BTW, what is $\tilde{h}_i$?} \ethan{CIC is known to CFD people; I forgot to revise the notation after moving the interpolation schemes to the appendix. will fix it}

\textbf{Multiscale Decoder Module. } We apply a linear projection $P\in\mathbb{R}^{2H\times H}$ and then reuse the GNS architecture from Sec~\ref{sec:gns} with $L$ layers using an independent set of parameters. In essence, we fuse global and local features for each particle for reconsideration in the local scale, which greatly improves the receptive field of each particle without significant compute overhead. We denote its output $\bh_i^\uparrow$. We then use an MLP $\varphi_\text{dec} : \mathbb{R}^H \rightarrow \mathbb{R}^d$ to yield an acceleration prediction $\hat{\ddot{\bx}}_i = \varphi_\text{dec}(\bh_i^\uparrow)$ for $i \in \llbracket 1, N \rrbracket$.

\subsection{Analysis of Information Propagation}

We further justify the importance of modeling dynamics globally, by considering the fact that advection propagates at the speed of sound. For a graph neural network whose neighborhoods are defined by a radius $r \in \mathbb{R}$, the maximum distance by which information can propagate for each time step is $r \times L$, where $L$ is the number of layers, therefore the corresponding propagation speed is upper bounded by $rL / \Delta_t$. 

In order to increase propagation speed while keeping the GNN structure, one must either increase neighborhood radius, temporal fidelity, or number of message passing steps. However, all of these methods have major limitations: increasing the neighborhood radius inflates graph connectivity and hence memory costs by $\mathcal{O}(R^2)$; increasing temporal fidelity extends inference time by $\mathcal{O}(\frac{1}{\Delta_t})$, and may cause rollout instability; increasing the number of message passing steps also extends inference/training time by $\mathcal{O}(L)$, and may induce over-smoothing. 
\ours, on the other hand, bridges this gap efficiently by directly learning global dynamics, relying on a separate Eulerian projection. We characterize the time complexity of \ours with the following theorem. 
\begin{theorem} \label{thm: CNN layer count}
Given GNN edge construction radius $r$, number of MP layers $L$, unit time step $\Delta_t$ and advection propagation speed $v$, the CNN module in \ours require a minimum of $\Omega (\log (v \Delta_t / rL))$ layers to properly model long-range global interactions. 
\end{theorem}
We leave the proof of this theorem to App~\ref{app: theory}. Note that as the number of particles $N$ increases, assuming we fix the Lebesgue measure of the ambient space, from the runtime and memory bottlenecks of GNN, $r$ must scale with $\mathcal{O} (1 / N^{1/d})$, which means the inference and training time of the \ours module scales roughly with $\mathcal{O}(\log N)$. This is a significant improvement compared with GNNs expanded to propagate information at scale with $v$, which suffers a complexity of $\mathcal{O} (L) = \mathcal{O} (1 / r) = \mathcal{O} (N^{1/d})$. 

\section{Experiments} \label{sec:experiments}
In our experiments, we use the datasets, evaluation metrics, and baselines implemented in LagrangeBench \citep{toshev2023lagrangebench}. We aim to show that \ours boosts performance on a variety of CFD tasks without much overhead in training or inference time compared to a GNS.

In our results, we provide performance metrics as well as timing information. Our rollout times are given as the mean among all rollout steps during evaluation. Our training time is given as the total training time. Both times are reported in A100 seconds.

\subsection{Settings}
\textbf{Datasets.}
In all of our experiments, we build upon LagrangeBench \citep{toshev2023lagrangebench}, a collection of seven Lagrangian benchmark datasets that cover a wide range of canonical incompressible flow configurations available under the MIT license. Each dataset is generated with a weakly–compressible Smoothed Particle Hydrodynamics (SPH) solver and, crucially for learning systems, every $100$th solver step is stored, turning the raw simulator output into a temporal coarse-graining task (i.e. the model must advance the flow $100$ physical time–steps at once). We elaborate on the main characteristics that are relevant for interpreting our quantitative results in App~\ref{app:datasets}.

\textbf{Computational Resources. } The experiments are done on Lambda's \texttt{gpu\_8x\_a100} cloud instances. The training time requirements for the above experiments are listed in the tables. In total, we estimate approximately 2500 A100 hours were used, including additional exploration not detailed here.

Our implementation details may be found in App~\ref{app:impl}. They closely reflect those found in LagrangeBench, as in our experience they achieve reasonable results.

\subsection{Main Results}
\begin{table}[t]
  \centering
  \caption{Metrics for each model on each dataset. Our results use the best checkpoint determined by $\text{MSE}_{20}$. The results are reported as $\mu\pm\sigma$, utilizing the entire test dataset. The distributions are generally positively skewed, and hence the errors are asymmetric. Due to outliers, the standard deviation may be larger than the mean.}
  \label{tab:results}
  \begin{tabular}{ccccccc}
    \toprule
    \textbf{Dataset} & \textbf{Model} &
      \multicolumn{1}{c}{$\text{MSE}_{20}$} &
      \multicolumn{1}{c}{$\text{MSE}_{E_{\text{kin}}}$} &
      \multicolumn{1}{c}{Sinkhorn} 
      & \multicolumn{1}{c}{Training time (s)}
      & \multicolumn{1}{c}{Inference time (s)} \\
\midrule
% ---------- 2D DATASETS -------------------------------------------------
\multirow{3}{*}{DAM-2D}
& GNS        & $\num{3.86e-05}\pm\num{2.86e-05}$ & $\num{1.59e-04}\pm\num{2.38e-04}$ & $\num{1.36e-05}\pm\num{1.71e-05}$ & \num{2.03e+04} & \num{1.15e-02} \\
& SEGNN      & $\num{5.04e-05}\pm\num{3.81e-05}$ & $\num{1.34e-04}\pm\num{1.98e-04}$ & $\num{2.38e-05}\pm\num{2.93e-05}$ & \num{4.01e+04} & \num{2.79e-02} \\
& \ours      & \textbf{$\num{1.55e-05}\pm\num{1.51e-05}$} & \textbf{$\num{2.18e-05}\pm\num{3.55e-05}$} & \textbf{$\num{2.82e-06}\pm\num{2.32e-06}$} & \num{2.10e+04} & \num{1.13e-02} \\
\midrule
\multirow{3}{*}{LDC-2D}
& GNS        & $\num{1.64e-05}\pm\num{2.32e-06}$ & $\num{4.64e-07}\pm\num{2.90e-07}$ & $\num{1.07e-06}\pm\num{2.76e-07}$ & \num{1.56e+04} & \num{8.72e-03} \\
& SEGNN      & $\num{1.91e-05}\pm\num{2.42e-06}$ & $\num{5.79e-07}\pm\num{4.20e-07}$ & $\num{1.72e-06}\pm\num{4.06e-07}$ & \num{2.73e+04} & \num{1.67e-02} \\
& \ours      & \textbf{$\num{1.45e-05}\pm\num{2.24e-06}$} & \textbf{$\num{3.81e-07}\pm\num{2.43e-07}$} & \textbf{$\num{5.07e-07}\pm\num{7.46e-08}$} & \num{1.78e+04} & \num{1.05e-02} \\
\midrule
\multirow{3}{*}{RPF-2D}
& GNS        & $\num{3.69e-06}\pm\num{7.31e-07}$ & $\num{2.81e-05}\pm\num{3.23e-05}$ & $\num{1.92e-07}\pm\num{6.90e-08}$ & \num{1.50e+04} & \num{1.03e-02} \\
& SEGNN      & $\num{3.51e-06}\pm\num{7.63e-07}$ & $\num{1.78e-05}\pm\num{1.93e-05}$ & $\num{2.74e-07}\pm\num{1.05e-07}$ & \num{2.25e+04} & \num{1.47e-02} \\
& \ours      & \textbf{$\num{1.54e-06}\pm\num{4.03e-07}$} & \textbf{$\num{2.39e-06}\pm\num{2.45e-06}$} & \textbf{$\num{2.08e-08}\pm\num{4.50e-09}$} & \num{1.68e+04} & \num{1.09e-02} \\
\midrule
\multirow{3}{*}{TGV-2D}
& GNS        & $\num{6.74e-06}\pm\num{9.75e-06}$ & $\num{4.64e-07}\pm\num{1.16e-06}$ & $\num{4.39e-07}\pm\num{5.57e-07}$ & \num{1.50e+04} & \num{8.63e-03} \\
& SEGNN      & $\num{4.40e-06}\pm\num{6.89e-06}$ & $\num{3.95e-07}\pm\num{1.09e-06}$ & $\num{1.85e-07}\pm\num{2.43e-07}$ & \num{2.19e+04} & \num{1.22e-02} \\
& \ours      & \textbf{$\num{3.81e-06}\pm\num{5.62e-06}$} & \textbf{$\num{2.90e-07}\pm\num{7.68e-07}$} & \textbf{$\num{1.05e-07}\pm\num{8.74e-08}$} & \num{1.61e+04} & \num{8.88e-03} \\
% ---------- 3D DATASETS -------------------------------------------------
\midrule
\multirow{3}{*}{LDC-3D}
& GNS        & $\num{4.15e-05}\pm\num{2.98e-06}$ & $\num{1.86e-08}\pm\num{1.53e-08}$ & $\num{4.82e-07}\pm\num{1.91e-07}$ & \num{3.79e+04} & \num{2.18e-02} \\
& SEGNN      & $\num{4.18e-05}\pm\num{3.00e-06}$ & $\num{4.00e-08}\pm\num{2.67e-08}$ & \textbf{$\num{2.64e-07}\pm\num{9.79e-08}$} & \num{9.02e+04} & \num{6.53e-02} \\
& \ours      & \textbf{$\num{3.86e-05}\pm\num{2.84e-06}$} & \textbf{$\num{1.55e-08}\pm\num{9.02e-09}$} & $\num{2.69e-07}\pm\num{9.87e-08}$ & \num{5.46e+04} & \num{2.64e-02} \\
\midrule
\multirow{3}{*}{RPF-3D}
& GNS        & $\num{2.08e-05}\pm\num{1.73e-06}$ & $\num{1.91e-06}\pm\num{1.89e-06}$ & $\num{2.15e-07}\pm\num{4.69e-08}$ & \num{2.75e+04} & \num{1.63e-02} \\
& SEGNN      & \textbf{$\num{1.64e-05}\pm\num{1.47e-06}$} & \textbf{$\num{1.34e-06}\pm\num{1.51e-06}$} & $\num{2.53e-07}\pm\num{8.97e-08}$ & \num{5.95e+04} & \num{4.30e-02} \\
& \ours      & $\num{1.95e-05}\pm\num{1.80e-06}$ & $\num{1.58e-06}\pm\num{1.51e-06}$ & \textbf{$\num{1.33e-07}\pm\num{2.24e-08}$} & \num{4.90e+04} & \num{2.23e-02} \\
\midrule
\multirow{3}{*}{TGV-3D}
& GNS        & $\num{7.17e-03}\pm\num{7.00e-03}$ & $\num{7.87e-02}\pm\num{8.05e-02}$ & $\num{8.78e-05}\pm\num{8.33e-05}$ & \num{2.87e+04} & \num{1.92e-02} \\
& SEGNN      & $\num{8.26e-03}\pm\num{1.77e-02}$ & \textbf{$\num{2.21e-03}\pm\num{4.42e-03}$} & $\num{1.73e-04}\pm\num{2.49e-04}$ & \num{6.40e+04} & \num{5.90e-02} \\
& \ours      & \textbf{$\num{6.10e-03}\pm\num{5.96e-03}$} & $\num{5.73e-02}\pm\num{5.43e-02}$ & \textbf{$\num{2.11e-05}\pm\num{1.60e-05}$} & \num{5.12e+04} & \num{2.11e-02} \\
    \bottomrule
  \end{tabular}
\end{table}

Using a GNS variant of \ours, on average we show a 57\% improvement in accuracy compared to GNS, while only utilizing 13\% more inference time and 31\% more training time. Compared to SEGNN, \ours on average achieves a 49\% improvement in accuracy, while also obtaining a 48\% decrease in inference time and a 30\% decrease in training time. The metrics are recorded in Table \ref{tab:results}. We elaborate on SEGNN's performance in Sec~\ref{sec:disc}, but we note here that SEGNN's inductive bias empirically seems to be helpful only on highly symmetrical and idealized scenarios, while also being significantly more expensive than GNS or \ours.

\begin{figure}[h]
  \centering                  % centre the whole block
  % ----------- First panel -----------
  \begin{subfigure}[t]{0.32\textwidth} % 3 × 0.32 < 1 ⇒ room for spacing
    \includegraphics[width=\linewidth]{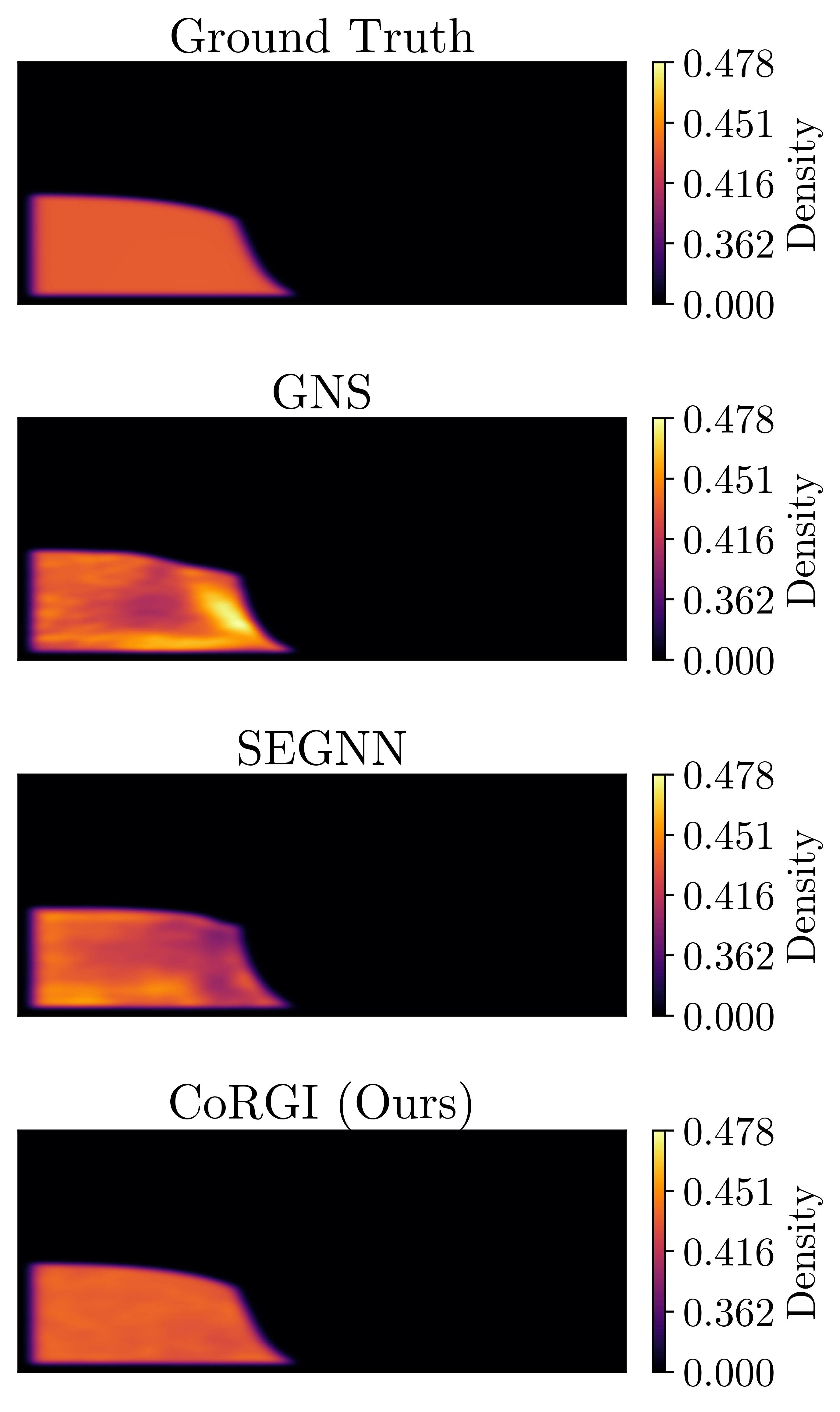}
    \caption{$t = 25$}
    % \label{fig:panelA}
  \end{subfigure}
  \hfill
  \begin{subfigure}[t]{0.32\textwidth}
    \includegraphics[width=\linewidth]{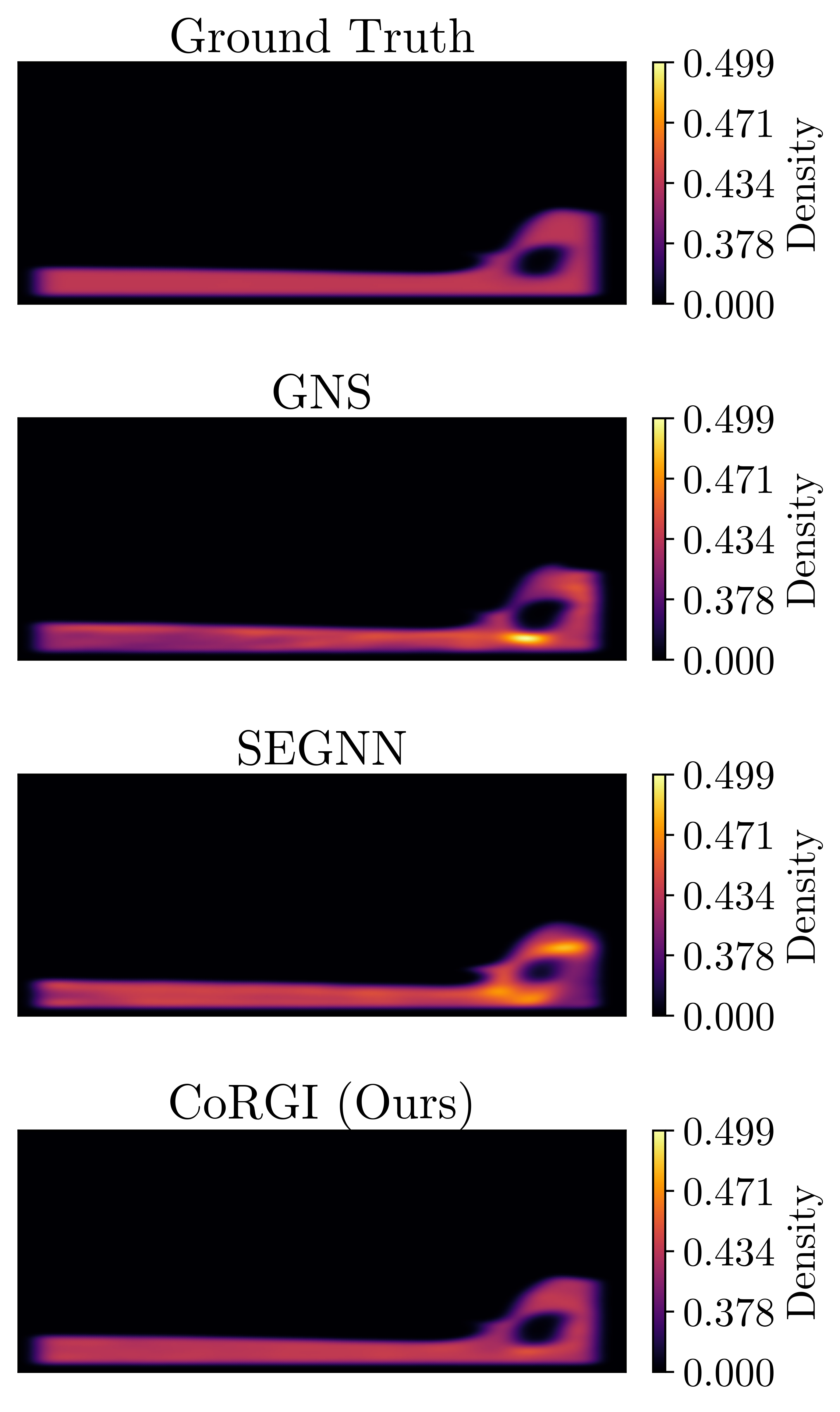}
    \caption{$t = 207$}
  \end{subfigure}
  \hfill
  % ----------- Second panel ----------
  \begin{subfigure}[t]{0.32\textwidth}
    \includegraphics[width=\linewidth]{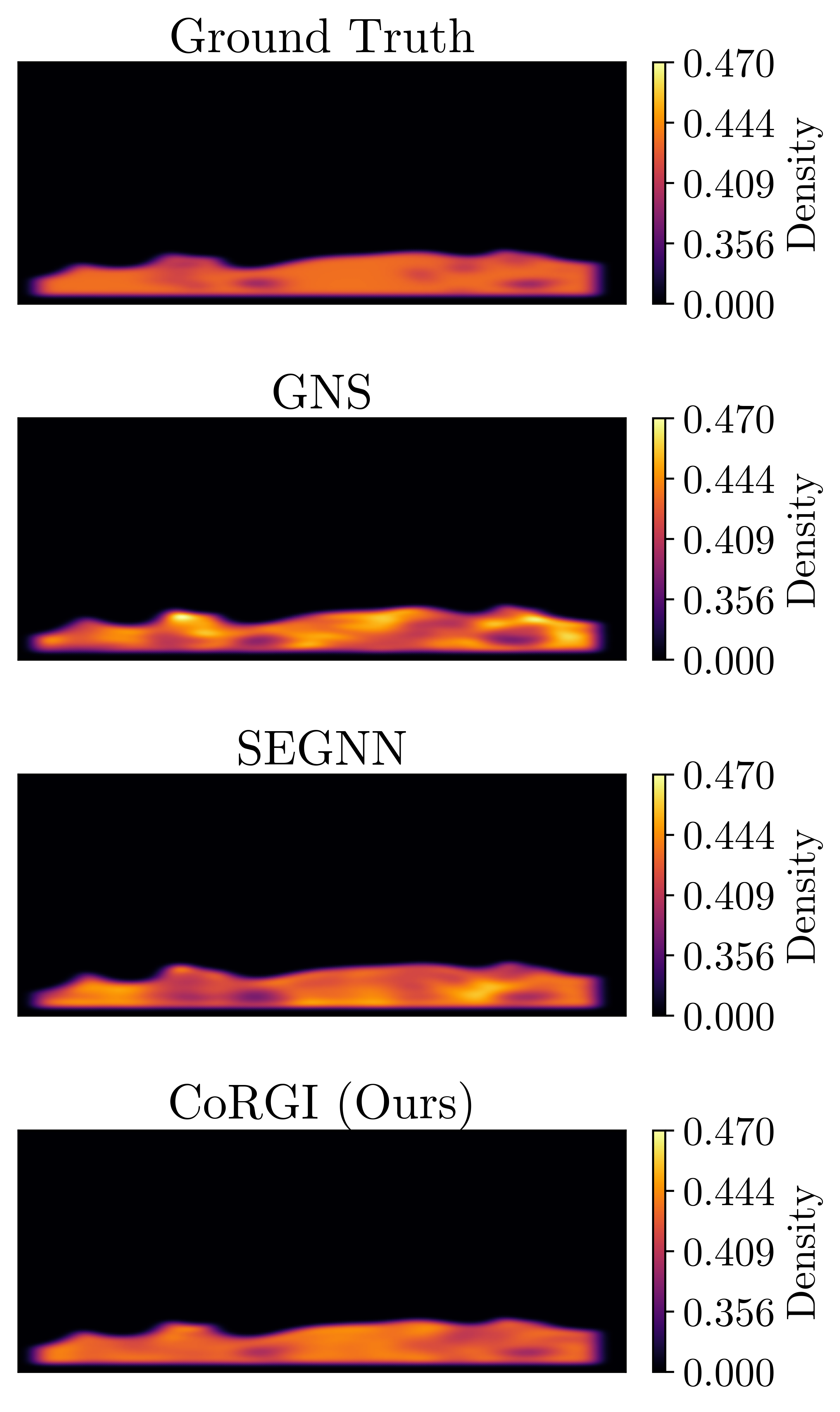}
    \caption{$t = 363$}
    % \label{fig:panelB}
  \end{subfigure}

  \caption{Kernel density estimation on DAM-2D ($t \in \llbracket 0, 400\rrbracket$). Above, uniform coloring indicates adherence to fluid incompressibility. Qualitatively, \ours maintains lower density variance under turbulent regimes.}
  \label{fig:dam}
\end{figure}

\begin{figure}[b]
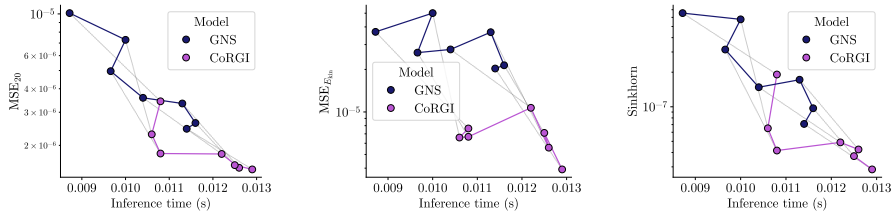

% \vspace{-0.1}
  \centering                  % centre the whole block
  % ----------- First panel -----------
  \begin{subfigure}[t]{0.32\textwidth} % 3 × 0.32 < 1 ⇒ room for spacing
    \includesvg[width=\linewidth]{figures/mse20_vs_time.svg}
    % \caption{$\text{MSE}_{20}$}
    % \label{fig:panelA}
  \end{subfigure}
  \hfill                         % stretchable horizontal space
  % ----------- Second panel ----------
  \begin{subfigure}[t]{0.32\textwidth}
    \includesvg[width=\linewidth]{figures/mse_ekin_vs_time.svg}
    % \caption{Kinetic energy MSE}
    % \label{fig:panelB}
  \end{subfigure}
  \hfill
  % ----------- Third panel -----------
  \begin{subfigure}[t]{0.32\textwidth}
    \includesvg[width=\linewidth]{figures/sinkhorn_vs_time.svg}
    % \caption{Sinkhorn}
    % \label{fig:panelC}
  \end{subfigure}

  \caption{Plot of accuracy compared to inference time for GNS and \ours on RPF-2D. Dotted lines pair points based on number of message passing steps. We adjust our allocated time budget in the above plots by adding or removing GNN message passing layers. Due to noise, the relationship is not empirically monotonic, but still demonstrates a trend of \ours outperforming GNS for a given time budget.}
  \label{fig:pareto}
\end{figure}

In our experiments, we compare against two established GNNs: GNS \citep{sanchezgonzalez2020learning} and SEGNN \citep{brandstetter2022geometric}. We choose these two models because they represent two different classes of GNN: non-equivariant (GNS) and equivariant (SEGNN). These are also identified as state of the art in the literature \citep{toshev2024neural}, and to the authors' knowledge, remain so. We note that PaiNN \citep{schütt2021equivariant} and EGNN \citep{satorras2022enequivariant}, although both relatively efficient equivariant models (i.e. one without expensive Clebsch-Gordan tensor products), have been shown to be unsuitable for CFD \citep{toshev2023lagrangebench}. 

Qualitatively, we also see in Figure~\ref{fig:dam} that \ours overall maintains a much more uniform density than our baselines. Fluid incompressibility constraints (i.e. uniform density) are traditionally difficult to maintain in situations with high Reynolds numbers (i.e. turbulence), such as in dam break. We demonstrate that, even in highly turbulent situations, \ours provides strong performance in physical plausibility. As evident in Table~\ref{tab:results}, \ours performs an order of magnitude better compared to baselines in kinetic energy error and Sinkhorn divergence, both of which rely heavily on fluid incompressibility.

\subsection{Additional Experiments}
We focus on the RPF-2D dataset for our additional experiments, as it provides the clearest insight in where \ours's improvement originates. Detailed results are present in Appendix~\ref{app:abl}. In summary, we find that the improvement offered by \ours over GNS is robust to changes in number of message passing steps and graph connectivity. In other words, we expect that \ours will improve upon GNS regardless of the specific budget for its graphical component.

% \subsubsection{Convolution}
% We demonstrate the performance impact of adding convolutional layers, as well as a good base feature size to perform well.

% In Table~\ref{tab:abl-conv}, we show that adding additional convolutional layers generally improves performance at the cost of additional inference and training time. However, additional layers past 3 seem to provide diminishing returns. Moreover, we also demonstrate that we can use a similar number of latent channels as the GNS, i.e. 128, and achieve optimal or near-optimal performance.

% Intuitively, we wish for each cell in the global aggregation step to have a global receptive field. This is achieved when we have 3 convolutional blocks.

% \subsubsection{Impact of Number of GNN Layers}

Building upon this idea, in Fig~\ref{fig:pareto}, we illustrate the accuracy with regard to inference time for our metrics. Overall, \ours demonstrates an average accuracy improvement of 47\% given a time budget (chosen from the range in which both GNS and \ours reside) over GNS. The metrics may be found in Table~\ref{tab:abl-mp}.

\begin{table}[t]
\centering
\caption{Performance based on number of message passing steps. We show that \ours improves upon GNS regardless of the number of message passing steps. Moreover, \ours often represents a better use of inference time budget, i.e. providing better accuracy for the same inference time.}
\label{tab:abl-mp}
  \begin{tabular}{ccccccc}
  % --------------------------------

    \toprule
    \textbf{MP steps} & \textbf{Model} &
      \multicolumn{1}{c}{$\text{MSE}_{20}$} &
      \multicolumn{1}{c}{$\text{MSE}_{E_{\text{kin}}}$} &
      \multicolumn{1}{c}{Sinkhorn} &
      \multicolumn{1}{c}{Training time (s)} &
      \multicolumn{1}{c}{Inference time (s)} \\
\midrule
\multirow{3}{*}{4}
& GNS & $\num{1.01e-05}\pm\num{1.71e-06}$ & $\num{3.69e-05}\pm\num{4.03e-05}$ & $\num{6.54e-07}\pm\num{1.70e-07}$ & $\num{1.14e+04}$ & $\num{8.72e-03}$ \\
& SEGNN & $\num{1.17e-05}\pm\num{1.99e-06}$ & $\num{3.55e-05}\pm\num{3.86e-05}$ & $\num{7.89e-07}\pm\num{2.67e-07}$ & $\num{1.36e+04}$ & $\num{8.99e-03}$ \\
& \ours & \textbf{$\num{3.43e-06}\pm\num{6.22e-07}$} & \textbf{$\num{7.64e-06}\pm\num{8.50e-06}$} & \textbf{$\num{1.91e-07}\pm\num{5.17e-08}$} & $\num{1.54e+04}$ & $\num{1.08e-02}$ \\
\midrule
\multirow{3}{*}{6}
& GNS & $\num{7.28e-06}\pm\num{1.47e-06}$ & $\num{5.01e-05}\pm\num{5.47e-05}$ & $\num{5.77e-07}\pm\num{1.90e-07}$ & $\num{1.35e+04}$ & $\num{1.00e-02}$ \\
& SEGNN & $\num{7.02e-06}\pm\num{1.38e-06}$ & $\num{2.68e-05}\pm\num{2.95e-05}$ & $\num{5.18e-07}\pm\num{2.02e-07}$ & $\num{1.73e+04}$ & $\num{1.14e-02}$ \\
& \ours & \textbf{$\num{2.29e-06}\pm\num{5.19e-07}$} & \textbf{$\num{6.59e-06}\pm\num{7.46e-06}$} & \textbf{$\num{6.50e-08}\pm\num{2.07e-08}$} & $\num{1.60e+04}$ & $\num{1.06e-02}$ \\
\midrule
\multirow{3}{*}{8}
& GNS & $\num{4.96e-06}\pm\num{9.93e-07}$ & $\num{2.63e-05}\pm\num{2.90e-05}$ & $\num{3.15e-07}\pm\num{9.45e-08}$ & $\num{1.40e+04}$ & $\num{9.66e-03}$ \\
& SEGNN & $\num{4.83e-06}\pm\num{9.78e-07}$ & $\num{1.97e-05}\pm\num{2.11e-05}$ & $\num{3.80e-07}\pm\num{1.48e-07}$ & $\num{2.07e+04}$ & $\num{1.35e-02}$ \\
& \ours & \textbf{$\num{1.81e-06}\pm\num{4.40e-07}$} & \textbf{$\num{6.67e-06}\pm\num{7.47e-06}$} & \textbf{$\num{4.16e-08}\pm\num{1.39e-08}$} & $\num{1.64e+04}$ & $\num{1.08e-02}$ \\
\midrule
\multirow{3}{*}{10}
& GNS & $\num{3.58e-06}\pm\num{8.29e-07}$ & $\num{2.77e-05}\pm\num{3.16e-05}$ & $\num{1.48e-07}\pm\num{5.74e-08}$ & $\num{1.56e+04}$ & $\num{1.04e-02}$ \\
& SEGNN & $\num{3.40e-06}\pm\num{7.32e-07}$ & $\num{1.78e-05}\pm\num{1.98e-05}$ & $\num{3.23e-07}\pm\num{1.29e-07}$ & $\num{2.29e+04}$ & $\num{1.50e-02}$ \\
& \ours & \textbf{$\num{1.80e-06}\pm\num{3.77e-07}$} & \textbf{$\num{1.07e-05}\pm\num{1.34e-05}$} & \textbf{$\num{4.90e-08}\pm\num{1.37e-08}$} & $\num{1.82e+04}$ & $\num{1.22e-02}$ \\
\midrule
\multirow{3}{*}{12}
& GNS & $\num{3.34e-06}\pm\num{7.91e-07}$ & $\num{3.67e-05}\pm\num{3.94e-05}$ & $\num{1.72e-07}\pm\num{6.42e-08}$ & $\num{1.66e+04}$ & $\num{1.13e-02}$ \\
& SEGNN & $\num{2.47e-06}\pm\num{5.60e-07}$ & $\num{1.88e-05}\pm\num{2.17e-05}$ & $\num{2.05e-07}\pm\num{8.77e-08}$ & $\num{2.59e+04}$ & $\num{1.70e-02}$ \\
& \ours & \textbf{$\num{1.52e-06}\pm\num{4.44e-07}$} & \textbf{$\num{5.59e-06}\pm\num{6.48e-06}$} & \textbf{$\num{4.25e-08}\pm\num{2.07e-08}$} & $\num{1.99e+04}$ & $\num{1.26e-02}$ \\
\midrule
\multirow{3}{*}{14}
& GNS & $\num{2.63e-06}\pm\num{6.10e-07}$ & $\num{2.14e-05}\pm\num{2.35e-05}$ & $\num{9.71e-08}\pm\num{3.59e-08}$ & $\num{1.78e+04}$ & $\num{1.16e-02}$ \\
& SEGNN & $\num{2.05e-06}\pm\num{4.45e-07}$ & $\num{1.13e-05}\pm\num{1.23e-05}$ & $\num{1.61e-07}\pm\num{7.33e-08}$ & $\num{2.96e+04}$ & $\num{1.92e-02}$ \\
& \ours & \textbf{$\num{1.57e-06}\pm\num{3.18e-07}$} & \textbf{$\num{7.11e-06}\pm\num{7.82e-06}$} & \textbf{$\num{3.72e-08}\pm\num{1.18e-08}$} & $\num{2.01e+04}$ & $\num{1.25e-02}$ \\
\midrule
\multirow{3}{*}{16}
& GNS & $\num{2.45e-06}\pm\num{5.48e-07}$ & $\num{2.03e-05}\pm\num{2.20e-05}$ & $\num{7.09e-08}\pm\num{2.67e-08}$ & $\num{1.84e+04}$ & $\num{1.14e-02}$ \\
& SEGNN & $\num{1.66e-06}\pm\num{3.46e-07}$ & $\num{9.40e-06}\pm\num{1.05e-05}$ & $\num{1.29e-07}\pm\num{6.03e-08}$ & $\num{3.37e+04}$ & $\num{2.16e-02}$ \\
& \ours & \textbf{$\num{1.49e-06}\pm\num{2.85e-07}$} & \textbf{$\num{3.92e-06}\pm\num{4.72e-06}$} & \textbf{$\num{2.85e-08}\pm\num{9.06e-09}$} & $\num{2.05e+04}$ & $\num{1.29e-02}$ \\
% \midrule
% \multirow{3}{*}{18}
% & GNS & $\num{2.22e-06}\pm\num{4.79e-07}$ & $\num{1.95e-05}\pm\num{2.22e-05}$ & $\num{5.56e-08}\pm\num{1.92e-08}$ & $\num{2.02e+04}$ & $\num{1.25e-02}$ \\
% & SEGNN & $\num{1.49e-06}\pm\num{3.50e-07}$ & $\num{1.10e-05}\pm\num{1.22e-05}$ & $\num{1.10e-07}\pm\num{5.60e-08}$ & $\num{3.55e+04}$ & $\num{2.28e-02}$ \\
% & \ours & $\num{1.47e-06}\pm\num{2.92e-07}$ & $\num{5.31e-06}\pm\num{6.52e-06}$ & $\num{2.62e-08}\pm\num{8.71e-09}$ & $\num{2.25e+04}$ & $\num{1.42e-02}$ \\
% \midrule
% \multirow{3}{*}{20}
% & GNS & $\num{1.97e-06}\pm\num{4.12e-07}$ & $\num{1.72e-05}\pm\num{1.83e-05}$ & $\num{3.47e-08}\pm\num{1.05e-08}$ & $\num{2.02e+04}$ & $\num{1.20e-02}$ \\
% & SEGNN & $\num{1.38e-06}\pm\num{3.23e-07}$ & $\num{1.13e-05}\pm\num{1.15e-05}$ & $\num{8.17e-08}\pm\num{4.13e-08}$ & $\num{3.84e+04}$ & $\num{2.46e-02}$ \\
% & \ours & $\num{1.46e-06}\pm\num{2.82e-07}$ & $\num{4.87e-06}\pm\num{5.46e-06}$ & $\num{2.99e-08}\pm\num{9.36e-09}$ & $\num{2.13e+04}$ & $\num{1.33e-02}$ \\
    \bottomrule
  \end{tabular}
\end{table}

\subsection{Discussion}
\label{sec:disc}

% \ethan{Actually I am not too convinced by the scaling of $r$. Firstly this relies on the assumption that the area/volume is the same and secondly wouldn't it be $O(1/n^2)$? Also maybe it would be better to use "measure", volume seems to be 3D

% I am not sure what you mean, like with enough message passing to cover everything? it should be parallelized over degree i think, yeah O(N) is probably also parallelized, that's why I said O(L) above…

% yeah probably}

% \ethan{yeah, i'm gonna have to move some stuff to the appendix, like the implementation details, we'll see how much space there is after i refactor that part

% sgtm}

% \ethan{I feel like I don't have that strong of an intuition on it, my understanding is that we simultaneously want a specific voxel density while also having enough layers for a global receptive field. so number of voxels should scale with number of particles linearly, which means number of layers would scale logarithmically

% I think it makes sense
% }

Overall, our experiments show that \ours excels in its intended purpose: to capture global information. Table~\ref{tab:results} shows that \ours always performs the best in Sinkhorn divergence, except in LDC-3D where it is only 2\% above SEGNN.

Intuitively, aggregating information on a global scale allows a model to better learn global dynamics. While $\text{MSE}_{20}$ and $\text{MSE}_{E_\text{kin}}$ are primarily local errors (i.e. $\text{MSE}_{20}$ measures pointwise positional error and $\text{MSE}_{E_\text{kin}}$ can be interpreted as pointwise energy error), Sinkhorn divergence measures the distance between particle distributions themselves, and hence is global in nature. Thus, it is not necessarily expected for \ours to outperform every model in every metric.

Nonetheless, the experiments in which SEGNN showed better performance are somewhat irreflective of real-world performance. For instance, RPF-3D is far more symmetrical than the typical real-world prediction task, and also has a low Reynolds number, and hence, an equivariant model may naturally learn such dynamics better. However, in more complex scenarios such as LDC-3D, \ours retains superior accuracy. Given our performance on these types of datasets, we believe \ours to be better for real-world use, including irregular or turbulent flows, at a fraction of the cost of SEGNN.

We additionally demonstrate the ability of \ours to better utilize training and inference time budget, i.e. that it is more useful to focus on both local and global interactions rather than exclusively to deepen the GNN. However, in some cases, it is still useful to integrate an equivariant inductive bias. We intend to explore this incorporation in future work, and we discuss this and other limitations in Appendix~\ref{sec:lim}.

\section{Conclusion}
In this paper, we introduce Convolutional Residual Global Interactions (\ours), a lightweight augmentation on GNN-based simulators which incorporates global knowledge into GNN predictions for hydrodynamics. By expanding the receptive field to a global scale, we overcome the subsonic propagation speed of a traditional GNN, and enable directly modeling nonlocal effects such as advection. We reduce GNS's error by 57\% out of the box, and when restricted to the same inference time budget, still demonstrate a 47\% improvement in accuracy. \ours through our testing appears as a highly promising and efficient architecture for computational fluid dynamics compared to existing arts in neural emulation.

% \section*{Reproducibility Statement}
% All required settings to reproduce experiments are detailed in the paper. The dataset used is described in App~\ref{app:datasets}, and our hyperparameters are detailed in App~\ref{app:impl}. All other settings are included as part of the relevant figures or tables. Code is provided as supplementary material.

\bibliography{iclr2026_conference}
\bibliographystyle{iclr2026_conference}

\appendix
\section{Theoretical Analysis} \label{app: theory}
% https://en.wikipedia.org/wiki/Courant%E2%80%93Friedrichs%E2%80%93Lewy_condition
In this section, we provide some further discussions to support the necessity of \ours, and prove Theorem~\ref{thm: CNN layer count}. The basis of our analysis is the Courant-Friedrich-Lewy condition \citep{courant1967partial}: 
\begin{equation} \label{eq: propagation speed}
\frac{v \Delta_t}{\Delta_{x}} \le 1, 
\end{equation}
where $v \in \mathbb{R}$ is the speed of sound, $\Delta_{x} \in \mathbb{R}$ is the length traveled per step (i.e. receptive field), and $\Delta_t \in \mathbb{R}$ is the length of a time step. We use this as a benchmark for the required speed of information propagation within the model.

\subsection{GNN Message Passing is Too Slow}
Interpreted in our setting, we can impose an upper bound on $v \Delta_t$ as $||b_\text{max} - b_\text{min}||_2$ where $b_\text{max}$ and $b_\text{min}$ are the maximal and minimal coordinates. We impose such an upper bound since this is the furthest distance data may travel in between time frames. In our analyses, we set the speed of sound to exactly this upper bound, as we interpret the speed of sound as some arbitrarily large value. It is evident that GNNs such as GNS or SEGNN will violate this condition with few message passing steps in this worst-case setting. Given a connectivity radius $r \in \mathbb{R}$ and $L \in \mathbb{N}$ message passing steps, we have from the analyses below that $||b_\text{max} - b_\text{min}||_2 / (rL)$ must be less than $1$ for effective information exchange. However, in typical settings, e.g. in LagrangeBench, the baseline GNNs yield Courant numbers much larger than $1$:

% I'm a bit confused on why we need to set speed = |max-min|? So the values in the table are computed with |max-min|? Kk that makes sense. 
% assume that the speed of sound is really fast, and also that it doesn't really make a difference whether it goes faster than that within a single frame idk yes

% Do you think it's good to separate this into two subsections? proof and table ; yep also I think they are pretty distinct sections
% maybe, i guess having a whole page without that kind of organization might be suboptimal agreed, maybe i should move it to the datasets section

\begin{table}[H]
  \caption{LagrangeBench Courant numbers (with $L = 10$)}
  \label{tab:phys-char}
  \centering
\begin{tabular}{ccccccc}
  \toprule
  \textbf{Dataset} & 
    \multicolumn{1}{c}{Dimensions} &
    \multicolumn{1}{c}{Connectivity radius} &
    \multicolumn{1}{c}{Courant number} \\
    \midrule
    DAM-2D & $5.586 \times 2.22$ & 0.029 & 20.73 \\
    LDC-2D & $1.12 \times 1.12$ & 0.029 & 5.46 \\
    RPF-2D & $1.0 \times 2.0$ & 0.036 & 6.21 \\
    TGV-2D & $1.0 \times 1.0$ & 0.029 & 4.88 \\
    LDC-3D & $1.25 \times 1.25 \times 0.5$ & 0.06 & 3.06 \\
    RPF-3D & $1.0 \times 2.0 \times 0.5$ & 0.072 & 3.18 \\
    TGV-3D & $2\pi \times 2\pi \times 2\pi$ & 0.46 & 2.37 \\
    \bottomrule
  \end{tabular}
\end{table}

\subsection{Proof of Theorem~\ref{thm: CNN layer count}: \ours Enables Fast Information Propagation}
\begin{proof}
First, we consider the furthest distance information can travel with $L$ layers of GNN message passing (MP). 
% More precisely, 
For a starting node $v \in \cV$, we denote nodes that are of degree at most $d$ (i.e., $d$-hop) from $v$ as $\cN_d(v)$, where note that $\cN_1(v)=\{ v' |(v, v') \in \cE \}$ are simply the neighbors of $v$. Then with $L$ MP layers, the information in $v$ can only propagate to nodes within $\cN_L(v)$. Since the graph is constructed with radius $r$, the maximum distance of this propagation is $rL$. 

Next, from \eqref{eq: propagation speed} we have the length traveled per step is $\Delta_x \geq v \Delta_t$, and so when $v \Delta_t > rL$, there is a sizable gap between the information propagation distance of the GNN and the minimum required distance per step, which \ours seeks to address. For our model, assume the grid constructed for CNNs at the highest resolution level has unit length $d_{\min}$ for each cell, while the lowest resolution level has unit length $d_{\max}$, then from the nature of CNN pooling, the number of layers must be at least $\Omega \big( \log (d_{\max} / d_{\min}) \big)$ (in the case of $2\times2$ pooling in our case, the logarithm takes base $2$). 

Now we analyze the required magnitudes for $d_{\min}$ and $d_{\max}$, respectively. For $d_{\min}$, since the grid scatter operation pools information over all nodes within a unit cell, the feature value for CNN represents the \textit{aggregated, averaged} information within the region of the unit cell. Therefore for each node within the cell to properly \textit{send} and \textit{receive} their more fine-grained individual information, respectively \textit{before} and \textit{after} the \ours module, we would require $d_{\min} = \cO (rL)$, or the unit cell length to not exceed the information propagation distance by the GNN in order of magnitude. For $d_{\max}$, we see the final layer of CNN allows interaction over maximum distance $C \times d_{\max}$, where $C$ is a constant associated with the CNN hyperparameters, such as kernel size and stride. Therefore we must have $d_{\max} = \Omega (v \Delta_t)$ to allow for the modeling of interaction on par with the Courant-Friedrich-Lewy condition \eqref{eq: propagation speed}. 

Combining everything, we finally have the number of CNN layers within \ours should be at least
$$ \Omega \big( \log (d_{\max} / d_{\min}) \big) = \Omega \big( \log (v \Delta_t / rL) \big). $$
\end{proof}

% For a \ours global convolution module of depth $d \in \mathbb{N}$, the receptive field of a corner voxel is $6 * 2^d - 2$, i.e. $\Delta_x \in \mathcal{O}(2^d)$. Based on the Courant-Friedrich-Lewy condition, we wish to achieve $\Delta_t v \le \Delta_x$. Rewriting this, we can see that $d \in \mathcal{O}(\log (\Delta_t v))$.

\section{Datasets} \label{app:datasets}
In all of our experiments, we build upon LagrangeBench \citep{toshev2023lagrangebench}, a collection of seven Lagrangian benchmark datasets that cover a wide range of canonical incompressible flow configurations available under the MIT license. Each dataset is generated with a weakly–compressible Smoothed Particle Hydrodynamics (SPH) solver and, crucially for learning systems, every 100th solver step is stored, turning the raw simulator output into a temporal coarse-graining task (i.e. the model must advance the flow 100 physical time–steps at once). Below we summarize the main characteristics that are relevant for interpreting our quantitative results.

\begin{table}[h]
  \caption{Physical characteristics}
  \label{tab:phys-char}
  \centering
  \begin{tabularx}{\textwidth}{lp{1.5cm}XX}
    \toprule
    % \multicolumn{2}{c}{Datasets}                   \\
    % \cmidrule(r){1-2}
    Geometry & Re & Boundary conditions & Key challenge \\
    \midrule
    Taylor-Green vortex & 100 (2D) \newline 50 (3D)  & fully periodic & rapid kinetic energy decay, symmetry preservation     \\
    Reverse Poiseuille flow     & 10 & periodic, body-force driven & stationary shear layers with spatially varying forcing      \\
    Lid-driven cavity     & 100       & moving lid, no-slip walls & accurate wall treatment, vortex formation  \\
    Dam break     & 40000       & free surface, walls & highly nonlinear free surface evolution  \\
    \bottomrule
  \end{tabularx}
\end{table}

The Reynolds numbers in the table correspond to the effective Re in the SPH solver settings. 3D variants use the same physics as their 2D counterparts, simply extruded in the out-of-plane direction (with periodicity enforced in $z$).

% We note that the dynamics between 2D and 3D datasets may be different for higher Reynold's numbers in turbulent regimes. In 3 dimensions, a phenomenon known as energy cascade transfers kinetic energy from large vortices to progressively smaller scales until it becomes thermal energy. In 2 dimensions, energy cascade facilitates the converse: smaller scale energy grows in size. Thus, in 3 dimensions, there are more vortical structures, whereas in 2 dimensions the behavior is more homogeneous.

\begin{table}[h]
  \caption{Numerical resolution and splits}
  \label{tab:num-res}
  \centering
  \begin{tabularx}{0.85\textwidth}{llp{4cm}ll}
    \toprule
    % \multicolumn{2}{c}{Datasets}                   \\
    % \cmidrule(r){1-2}
    Datasets & \# particles & Traj. length \newline (train / val / test) & $\Delta t$ & Convolution res. \\
    \midrule
    TGV-2D & 2500 & 126 / 63 / 63 & 0.04 & $32 \times 32$ \\
    RPF-2D & 3200 & 20000 / 10000 / 10000 & 0.04 & $32 \times 64$ \\
    LDC-2D & 2708 & 10000 / 5000 / 5000 & 0.04 & $32 \times 32$ \\
    DAM-2D & 5740 & 401 / 200 / 200 & 0.03 & $80 \times 32$ \\
    TGV-3D & 8000 & 61 / 31 / 31 & 0.5 & $32 \times 32 \times 32$ \\
    RPF-3D & 8000 & 10000 / 5000 / 5000 & 0.1 & $32 \times 64 \times 16$ \\
    LDC-3D & 8160 & 10000 / 5000 / 5000 & 0.09 & $40 \times 40 \times 16$ \\
    \bottomrule
  \end{tabularx}
\end{table}

For statistically stationary cases (RPF, LDC) the simulator is run until equilibrium and a single long trajectory is partitioned into the above splits; for transient cases (TGV, DAM) multiple independent realizations are generated.

The step size $\Delta t$ refers to the physical time between two stored frames (i.e. one ML training step equals 100 SPH sub-steps and corresponds to $\Delta t$).

The collection was chosen to expose learning algorithms to a broad span of flow phenomena: decay to rest, driven steady states, wall-bounded recirculation, and violent free-surface motion; and progressively harder geometric settings (from fully periodic boxes to complex wall topologies). Moreover, by subsampling at $100\times$ the solver step, the benchmark explicitly tests a model's ability to extrapolate in time, which is a key requirement for practical surrogate solvers in engineering applications.
\section{Implementation Details}
\label{app:impl}
\paragraph{Pipeline. }
In training and evaluation, we first isolate a subtrajectory of a suitable length in every trajectory we use. The length of this subtrajectory is determined by how long our historical inputs are and how many rollout steps we take. In our case, we use a history length of 6, and we do 20 rollout steps, as these are the defaults provided by LagrangeBench, and hence we aim for a trajectory length of 26. When initially predicting a trajectory, we start with 6 ground truth states, which are aggregated into suitable features (namely position and velocity, detailed in Sec~\ref{sec:gcon}) and given to the model. At each time step, we recompute the neighbor graph. Then, we drop the least recent frame and append the next frame generated by the model's prediction via symplectic Euler integration.

\paragraph{Learning Parameters. }
All models were assessed with exponential decay. The initial learning rate is \num{5e-4} and the final learning rate is \num{1e-6}. We use a decay rate of 0.1 and \num{1e5} decay steps. These are the same as in LagrangeBench. For GNS with and without \ours, we use a batch size of 4. In our experience, SEGNN had better performance with a batch size of 1 using the same learning rates, hence we report its results with the batch size of 1. All models used 10 message passing layers unless stated otherwise.

\paragraph{Hyperparameters. }
For our convolutional layers, we use a feature plan of $(128, 256, 512)$, as it has a relatively good performance with low overhead compared to GNS. We choose our resolution to ensure the receptive field of each cell is roughly global. We attempt to justify these choices in Appendix~\ref{app:abl}. The specific resolutions are given in Appendix~\ref{app:datasets}. In our main results, all of the models are assessed with 10 message passing layers, and graphical components all have a hidden dimension of 128; these are the defaults from LagrangeBench, and in our experience, offer reasonable performance. 

\paragraph{Metrics. } \label{sec:metr}
To assess how well a surrogate advances a particle system over many solver steps, we adopt three error measures: mean-squared position error, sinkhorn distance and kinetic energy MSE (See Appendix~\ref{app:metrics} for full description). All metrics are computed on rollouts of 20 steps and are averaged over every stored frame, particle and test trajectory, with the checkpoint chosen by the lowest validation $\text{MSE}_{20}$. When training, our loss is the MSE between the predicted positions and the ground truth predictions detailed in Appendix~\ref{sec:mse}.
\section{Evaluation Metrics} \label{app:metrics}
To assess how well a surrogate advances a particle system over many solver steps, we adopt exactly the three error measures proposed by LagrangeBench. All metrics are computed on full roll-outs of length $n$ steps (we report results for $n = 20$) and are averaged over every stored frame, particle and test trajectory; the quoted numbers are finally averaged over three independent training runs, with the checkpoint chosen by the lowest validation $\text{MSE}_{20}$.

\subsection{Mean-squared position error} \label{sec:mse}
We track the $\mathcal{L}_2$ error between predicted and ground truth particle positions, aggregated over an $n$-step rollout: \[\text{MSE}_n = \frac{1}{Nn} \sum_{k=1}^n \sum_{i=1}^N \lVert x_i^{(t+k)} - \hat{x}_i^{(t+k)} \rVert_2^2 \]

\subsection{Sinkhorn distance}
Pure position errors ignore that particles are interchangeable. We thus use the entropy-regularized optimal transport distance (i.e. Sinkhorn loss \citep{cuturi2013sinkhorn}) \[\mathcal{S}_\varepsilon (\mathbf{x}, \hat{\mathbf{x}}) = \min_{\Gamma \in \Pi(\mathbf{u},\mathbf{v})}{\langle\Gamma, C\rangle} + \varepsilon \text{KL}(\Gamma \lVert\mathbf{uv}^\intercal)\] where $C_{ij} = \lVert\mathbf{x}_i - \hat{\mathbf{x}}_i\rVert_2^2$ is a cost matrix and $\Pi(\mathbf{u}, \mathbf{v})$ is the set of couplings whose row and column sums match the source mass marginal distribution $\mathbf{u}$ and the target mass marginal distribution $\mathbf{v}$. In our case, each row and column in the coupling matrix must sum to $\frac{1}{N}$, i.e. the mass of a particle.

\subsection{Kinetic energy MSE}
To capture global physical consistency we compare the system's kinetic energy time series \[T(t) = \frac{1}{2} \sum_{i=1}^N m_i \lVert \dot{x}_i^{(t)}\rVert^2\] and report the mean-squared error between predicted and reference curves over the rollout window.

\subsection{Kernel-based metrics}
In order to assess the physical plausibility of \ours's improvements, we also employ some kernel-based metrics using the quintic spline \citep{morris1997modeling}.

Let the smoothing length $h$ be the average distance between particles. Then, the quintic spline is defined as $$W(r) = C_d h^{-d} (\max{(0, 3-\tfrac{r}{h})^5 - 6\max{(0, 2 - \tfrac{r}{h})}^3} + 15\max{(0, 1-\tfrac{r}{h}}))$$ where $C_2 = \tfrac{7}{478\pi}$ and $C_3 = \tfrac{1}{120\pi}$. Using $\mathbf{r}_{ij} \in\mathbb{R}^d$ to represent the displacement between two particles $i$ and $j$, we write the directional gradient $$\nabla W_{ij} = \frac{\mathbf{r}_{ij}}{||\mathbf{r}_{ij}||} \frac{d}{d||\mathbf{r}_{ij}||} W(||\mathbf{r}_{ij}||)$$

For divergence, we define the divergence at time step $t$ for particle $i$ to be $$\frac{\sum_{j} (\dot{\mathbf{x}}^t_{j} - \dot{\mathbf{x}}^t_{i}) \cdot \nabla W_{i j}}{\sum_{j} W(||\mathbf{x}_j - \mathbf{x}_i||)}$$

For vorticity, we perform a similar computation to divergence for the 3 dimensional case: $$\frac{\sum_{j} (\dot{\mathbf{x}}^t_{j} - \dot{\mathbf{x}}^t_{i}) \times \nabla W_{i j}}{\sum_{j} W(||\mathbf{x}_j - \mathbf{x}_i||)}$$ For the 2 dimensional case, embed the vector space in 3 dimensions, setting the third coordinate to 0.

For both divergence and vorticity, the reported metrics are the MSE over all particles and time steps, where error is defined to be the $\mathcal{L}_2$ distance.

\begin{table}[H]
  \centering
  \caption{Kernel based metrics.}
  \label{tab:krnl}
  \begin{tabular}{ccccccc}
    \toprule
    \textbf{Dataset} & \textbf{Model} &
      \multicolumn{1}{c}{Divergence error} &
      \multicolumn{1}{c}{Vorticity error} \\
\midrule
\multirow{2}{*}{DAM-2D}
& GNS & $\num{1.20e+00}\pm\num{1.01e+00}$ & $\num{2.73e+00}\pm\num{2.76e+00}$ \\
& \ours & $\num{1.04e+00}\pm\num{9.93e-01}$ & $\num{2.44e+00}\pm\num{2.80e+00}$ \\
\midrule
\multirow{2}{*}{LDC-2D}
& GNS & $\num{7.76e-01}\pm\num{4.76e-02}$ & $\num{1.64e+00}\pm\num{1.31e-01}$ \\
& \ours & $\num{6.54e-01}\pm\num{4.19e-02}$ & $\num{1.55e+00}\pm\num{1.23e-01}$ \\
\midrule
\multirow{2}{*}{RPF-2D}
& GNS & $\num{4.11e-02}\pm\num{2.69e-03}$ & $\num{9.51e-02}\pm\num{9.80e-03}$ \\
& \ours & $\num{2.97e-02}\pm\num{3.19e-03}$ & $\num{7.41e-02}\pm\num{1.10e-02}$ \\
\midrule
\multirow{2}{*}{TGV-2D}
& GNS & $\num{7.93e-02}\pm\num{1.23e-01}$ & $\num{2.17e-01}\pm\num{3.39e-01}$ \\
& \ours & $\num{8.38e-02}\pm\num{1.27e-01}$ & $\num{2.23e-01}\pm\num{3.44e-01}$ \\
\midrule
\multirow{2}{*}{LDC-3D}
& GNS & $\num{1.16e-01}\pm\num{3.63e-03}$ & $\num{3.99e-01}\pm\num{1.44e-02}$ \\
& \ours & $\num{1.12e-01}\pm\num{4.03e-03}$ & $\num{3.83e-01}\pm\num{1.54e-02}$ \\
\midrule
\multirow{2}{*}{RPF-3D}
& GNS & $\num{4.50e-02}\pm\num{1.35e-03}$ & $\num{1.16e-01}\pm\num{3.66e-03}$ \\
& \ours & $\num{4.39e-02}\pm\num{1.47e-03}$ & $\num{1.13e-01}\pm\num{3.96e-03}$ \\
\midrule
\multirow{2}{*}{TGV-3D}
& GNS & $\num{4.40e-03}\pm\num{4.35e-03}$ & $\num{1.66e-02}\pm\num{1.63e-02}$ \\
& \ours & $\num{2.82e-03}\pm\num{2.78e-03}$ & $\num{1.15e-02}\pm\num{1.13e-02}$ \\
\bottomrule
\end{tabular}
\end{table}

Based on our divergence errors, \ours typically respects fluid incompressibility better than GNS. Moreover, based on the vorticity errors, our results in Table~\ref{tab:results} were achieved without arbitrary dampening of angular momentum. Hence, \ours retains physical realism comparable or superior to GNS while offering superior pointwise accuracy.
\section{Additional Experiments}
\label{app:abl}

\subsection{Ablation}
We wish to show that our intuitions regarding the architecture of \ours are empirically justified. Hence, we conduct several other experiments in order to gain intuition on the performance characteristics of \ours. These experiments are mostly done with RPF-2D, with another dataset used only if otherwise stated.

\subsubsection{Skip Connections}
We assess which skip connections are most responsible for \ours's performance based on convolution depth. We denote the presence of a skip with either a $\top$ (there is a skip connection) or a $\bot$ (there is not a skip connection).

\begin{table}[H]
\centering
\caption{Performance based on skip connections}
\label{tab:abl-skip}
\begin{tabular}{ccccccc}
  \toprule
  \textbf{Skips} &
    \multicolumn{1}{c}{$\text{MSE}_{20}$} &
    \multicolumn{1}{c}{$\text{MSE}_{E_{\text{kin}}}$} &
    \multicolumn{1}{c}{Sinkhorn} &
    \multicolumn{1}{c}{Training time} &
    \multicolumn{1}{c}{Inference time} \\
\midrule
$\bot,\bot,\bot$ & $\num{1.76e-06}\pm\num{4.44e-07}$ & $\num{1.18e-05}\pm\num{1.34e-05}$ & $\num{8.23e-08}\pm\num{3.64e-08}$ & $\num{1.84e+04}$ & $\num{1.21e-02}$ \\
\midrule
$\bot,\bot,\top$ & $\num{1.60e-06}\pm\num{4.07e-07}$ & $\num{4.67e-06}\pm\num{4.56e-06}$ & $\num{3.34e-08}\pm\num{1.26e-08}$ & $\num{1.71e+04}$ & $\num{1.10e-02}$ \\
\midrule
$\bot,\top,\bot$ & $\num{1.68e-06}\pm\num{4.41e-07}$ & $\num{8.54e-06}\pm\num{1.03e-05}$ & $\num{4.41e-08}\pm\num{1.83e-08}$ & $\num{1.78e+04}$ & $\num{1.13e-02}$ \\
\midrule
$\bot,\top,\top$ & $\num{1.55e-06}\pm\num{4.28e-07}$ & $\num{6.21e-06}\pm\num{7.10e-06}$ & $\num{3.11e-08}\pm\num{1.20e-08}$ & $\num{1.77e+04}$ & $\num{1.13e-02}$ \\
\midrule
$\top,\bot,\bot$ & $\num{2.08e-06}\pm\num{4.56e-07}$ & $\num{1.78e-05}\pm\num{1.95e-05}$ & $\num{8.91e-08}\pm\num{3.94e-08}$ & $\num{1.83e+04}$ & $\num{1.21e-02}$ \\
\midrule
$\top,\bot,\top$ & $\num{1.69e-06}\pm\num{4.41e-07}$ & $\num{5.51e-06}\pm\num{5.70e-06}$ & $\num{4.05e-08}\pm\num{1.48e-08}$ & $\num{1.73e+04}$ & $\num{1.11e-02}$ \\
\midrule
$\top,\top,\bot$ & $\num{2.27e-06}\pm\num{4.14e-07}$ & $\num{1.63e-05}\pm\num{1.76e-05}$ & $\num{7.53e-08}\pm\num{2.50e-08}$ & $\num{1.86e+04}$ & $\num{1.24e-02}$ \\
\midrule
$\top,\top,\top$ & $\num{1.58e-06}\pm\num{4.10e-07}$ & $\num{4.55e-06}\pm\num{4.71e-06}$ & $\num{4.41e-08}\pm\num{1.47e-08}$ & $\num{1.92e+04}$ & $\num{1.25e-02}$ \\
  \bottomrule
\end{tabular}
\end{table}

In Table~\ref{tab:abl-skip}, we assess the importance of skip connections at each depth. From left to right, our skip notation denotes the presence of skips from the finest to coarsest resolutions in a \ours with a global convolution module of depth 3. In this table, it represents both the A and B skip connections of a particular depth, as denoted in Fig~\ref{fig:arch}.

The data show that the skip connections at the coarsest resolution provide the most significant improvement on performance, followed by the middle layer. Curiously, having a skip connection for the finest layer actually hurts performance. We hypothesize this degradation to be caused by the additional learning inertia: in the finest resolution, the skip connection concatenates information both before and after only one double convolution.

We also note that the training and inference times are not significantly affected by the presence of these skip connections.

\begin{table}[H]
\centering
\caption{Performance based on skip connections}
\label{tab:abl-skip2}
\begin{tabular}{ccccccc}
  \toprule
  \textbf{Skips} &
    \multicolumn{1}{c}{$\text{MSE}_{20}$} &
    \multicolumn{1}{c}{$\text{MSE}_{E_{\text{kin}}}$} &
    \multicolumn{1}{c}{Sinkhorn} &
    \multicolumn{1}{c}{Training time} &
    \multicolumn{1}{c}{Inference time} \\
\midrule
$\bot,\bot$ & $\num{1.76e-06}\pm\num{4.44e-07}$ & $\num{1.18e-05}\pm\num{1.34e-05}$ & $\num{8.23e-08}\pm\num{3.64e-08}$ & $\num{1.84e+04}$ & $\num{1.21e-02}$ \\
\midrule
$\bot,\top$ & $\num{1.77e-06}\pm\num{3.58e-07}$ & $\num{1.11e-05}\pm\num{1.36e-05}$ & $\num{6.99e-08}\pm\num{2.21e-08}$ & $\num{1.56e+04}$ & $\num{9.35e-03}$ \\
\midrule
$\top,\bot$ & $\num{1.86e-06}\pm\num{3.73e-07}$ & $\num{9.43e-06}\pm\num{1.07e-05}$ & $\num{5.97e-08}\pm\num{2.08e-08}$ & $\num{1.61e+04}$ & $\num{9.42e-03}$ \\
\midrule
$\top,\top$ & $\num{1.58e-06}\pm\num{4.10e-07}$ & $\num{4.55e-06}\pm\num{4.71e-06}$ & $\num{4.41e-08}\pm\num{1.47e-08}$ & $\num{1.92e+04}$ & $\num{1.25e-02}$ \\
  \bottomrule
\end{tabular}
\end{table}

We additionally show the necessity of both sets of skip connections (i.e. both A and B in Fig~\ref{fig:arch}) in Table~\ref{tab:abl-skip2}. Although neither one provides much of a boost in accuracy on their own, we demonstrate that including both sets of skip connections is the most optimal, without a significant overhead in training or inference time.

\subsubsection{Convolution}
In Table~\ref{tab:abl-conv}, we show the importance of our global convolution module, in particular comparing versions with and without any convolution operations. For a depth of 0, we directly use a GNS as a baseline. Even after adding a single convolutional level, we see drastic improvements across all metrics. Adding on more depth generally improves performance, however we see there are diminishing returns past 3 resolution levels, as the receptive field is nearly global at that point, and hence we fulfill the conditions from App~\ref{app: theory}.

Moreover, we also assess a convolution-only architecture \citep{rochman-sharabi2025a} below in Table~\ref{tab:mpm}. We omit a comparison on the RPF-3D dataset due to technical difficulties. However, across all other datasets, \ours performs significantly better.

\begin{table}[H]
  \centering
  \caption{Performance using only a U-Net, i.e. NeuralMPM. As LagrangeBench computes the neighbor graph regardless of the model's dependency on it, and we conduct this experiment as an ablation rather as a standalone experiment, our reported inference and training times include the time to construct the graph.}
  \label{tab:mpm}
  \begin{tabular}{ccccccc}
    \toprule
    \textbf{Dataset} &
      \multicolumn{1}{c}{$\text{MSE}_{20}$} &
      \multicolumn{1}{c}{$\text{MSE}_{E_{\text{kin}}}$} &
      \multicolumn{1}{c}{Sinkhorn} 
      & \multicolumn{1}{c}{Training time (s)}
      & \multicolumn{1}{c}{Inference time (s)} \\
\midrule
DAM-2D & $\num{4.95e-05}\pm\num{5.91e-05}$ & $\num{6.48e-05}\pm\num{2.05e-04}$ & $\num{1.27e-05}\pm\num{2.01e-05}$ & $\num{1.41e+04}$ & $\num{2.01e-02}$ \\
\midrule
LDC-2D & $\num{2.57e-05}\pm\num{2.05e-06}$ & $\num{2.29e-06}\pm\num{1.49e-06}$ & $\num{3.13e-07}\pm\num{5.19e-08}$ & $\num{1.27e+04}$ & $\num{1.40e-02}$ \\
\midrule
RPF-2D & $\num{4.87e-05}\pm\num{3.58e-06}$ & $\num{4.72e-05}\pm\num{5.08e-05}$ & $\num{1.08e-07}\pm\num{3.19e-08}$ & $\num{1.40e+04}$ & $\num{2.31e-02}$ \\
\midrule
TGv-2D & $\num{1.87e-05}\pm\num{2.57e-05}$ & $\num{1.46e-06}\pm\num{3.11e-06}$ & $\num{-6.63e-09}\pm\num{6.10e-07}$ & $\num{1.40e+04}$ & $\num{2.08e-02}$ \\
\midrule
LDC-3D & $\num{1.03e-04}\pm\num{3.84e-06}$ & $\num{1.16e-07}\pm\num{5.04e-08}$ & $\num{5.23e-06}\pm\num{6.54e-07}$ & $\num{2.03e+04}$ & $\num{1.68e-02}$ \\
\midrule
TGV-3D & $\num{1.61e-02}\pm\num{1.49e-02}$ & $\num{9.43e-01}\pm\num{9.52e-01}$ & $\num{3.28e-04}\pm\num{2.71e-04}$ & $\num{1.48e+04}$ & $\num{2.07e-02}$ \\
\bottomrule
\end{tabular}
\end{table}

\subsubsection{History}
We additionally assess the impact of history length in model performance. Our motivation is that macroscopic physics is approximately deterministic, i.e. the future state of a system should be entirely determinable from a state and a set of physical laws. Nonetheless, in approximations, it may be helpful to include historical information in order to numerically approximate acceleration, jerk, etc.

\begin{figure}[H]
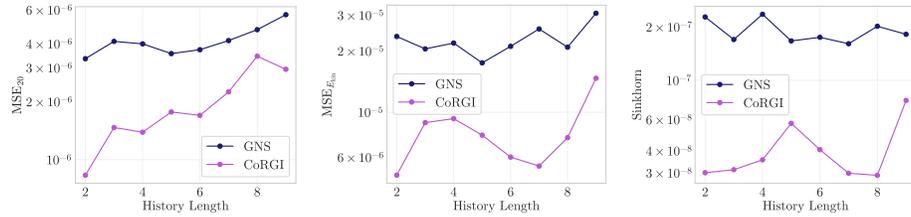

% \vspace{-0.1}
  \centering                  % centre the whole block
  % ----------- First panel -----------
  \begin{subfigure}[t]{0.32\textwidth} % 3 × 0.32 < 1 ⇒ room for spacing
    \includesvg[width=\linewidth]{figures/mse_20_vs_history.svg}
    % \caption{$\text{MSE}_{20}$}
    % \label{fig:panelA}
  \end{subfigure}
  \hfill                         % stretchable horizontal space
  % ----------- Second panel ----------
  \begin{subfigure}[t]{0.32\textwidth}
    \includesvg[width=\linewidth]{figures/mse_ekin_vs_history.svg}
    % \caption{Kinetic energy MSE}
    % \label{fig:panelB}
  \end{subfigure}
  \hfill
  % ----------- Third panel -----------
  \begin{subfigure}[t]{0.32\textwidth}
    \includesvg[width=\linewidth]{figures/sinkhorn_vs_history.svg}
    % \caption{Sinkhorn}
    % \label{fig:panelC}
  \end{subfigure}

  \caption{Plot of accuracy compared to history for GNS and \ours on RPF-2D.}
  % \label{fig:pareto}
\end{figure}

\begin{table}[H]
\centering
\caption{Performance based on history length}
\label{tab:abl-hist}
\begin{tabular}{ccccccc}
  \toprule
  \textbf{History} & \textbf{Model} &
    \multicolumn{1}{c}{$\text{MSE}_{20}$} &
    \multicolumn{1}{c}{$\text{MSE}_{E_{\text{kin}}}$} &
    \multicolumn{1}{c}{Sinkhorn} &
    \multicolumn{1}{c}{Training time} &
    \multicolumn{1}{c}{Inference time} \\
\midrule
\multirow{2}{*}{2}
& GNS & $\num{3.35e-06}\pm\num{8.33e-07}$ & $\num{2.36e-05}\pm\num{2.50e-05}$ & $\num{2.28e-07}\pm\num{7.71e-08}$ & $\num{1.56e+04}$ & $\num{1.07e-02}$ \\
& \ours & $\num{8.31e-07}\pm\num{2.22e-07}$ & $\num{4.91e-06}\pm\num{5.64e-06}$ & $\num{3.01e-08}\pm\num{9.68e-09}$ & $\num{1.88e+04}$ & $\num{1.22e-02}$ \\
\midrule
\multirow{2}{*}{3}
& GNS & $\num{4.12e-06}\pm\num{9.69e-07}$ & $\num{2.05e-05}\pm\num{2.42e-05}$ & $\num{1.70e-07}\pm\num{5.59e-08}$ & $\num{1.49e+04}$ & $\num{9.92e-03}$ \\
& \ours & $\num{1.47e-06}\pm\num{3.10e-07}$ & $\num{8.89e-06}\pm\num{9.98e-06}$ & $\num{3.13e-08}\pm\num{8.70e-09}$ & $\num{1.83e+04}$ & $\num{1.19e-02}$ \\
\midrule
\multirow{2}{*}{4}
& GNS & $\num{4.00e-06}\pm\num{8.87e-07}$ & $\num{2.19e-05}\pm\num{2.48e-05}$ & $\num{2.36e-07}\pm\num{7.91e-08}$ & $\num{1.53e+04}$ & $\num{1.01e-02}$ \\
& \ours & $\num{1.39e-06}\pm\num{2.97e-07}$ & $\num{9.31e-06}\pm\num{1.15e-05}$ & $\num{3.56e-08}\pm\num{1.12e-08}$ & $\num{1.71e+04}$ & $\num{1.09e-02}$ \\
\midrule
\multirow{2}{*}{5}
& GNS & $\num{3.56e-06}\pm\num{8.68e-07}$ & $\num{1.75e-05}\pm\num{2.03e-05}$ & $\num{1.67e-07}\pm\num{5.98e-08}$ & $\num{1.56e+04}$ & $\num{1.08e-02}$ \\
& \ours & $\num{1.77e-06}\pm\num{3.52e-07}$ & $\num{7.72e-06}\pm\num{8.78e-06}$ & $\num{5.72e-08}\pm\num{1.71e-08}$ & $\num{1.75e+04}$ & $\num{1.12e-02}$ \\
\midrule
\multirow{2}{*}{6}
& GNS & $\num{3.73e-06}\pm\num{7.99e-07}$ & $\num{2.11e-05}\pm\num{2.32e-05}$ & $\num{1.75e-07}\pm\num{6.39e-08}$ & $\num{1.59e+04}$ & $\num{1.08e-02}$ \\
& \ours & $\num{1.70e-06}\pm\num{3.55e-07}$ & $\num{6.02e-06}\pm\num{7.26e-06}$ & $\num{4.06e-08}\pm\num{1.12e-08}$ & $\num{1.73e+04}$ & $\num{1.10e-02}$ \\
\midrule
\multirow{2}{*}{7}
& GNS & $\num{4.16e-06}\pm\num{9.99e-07}$ & $\num{2.57e-05}\pm\num{2.76e-05}$ & $\num{1.61e-07}\pm\num{5.48e-08}$ & $\num{1.50e+04}$ & $\num{1.00e-02}$ \\
& \ours & $\num{2.25e-06}\pm\num{5.67e-07}$ & $\num{5.44e-06}\pm\num{5.91e-06}$ & $\num{2.99e-08}\pm\num{1.11e-08}$ & $\num{1.81e+04}$ & $\num{1.15e-02}$ \\
\midrule
\multirow{2}{*}{8}
& GNS & $\num{4.74e-06}\pm\num{8.94e-07}$ & $\num{2.09e-05}\pm\num{2.30e-05}$ & $\num{2.02e-07}\pm\num{7.49e-08}$ & $\num{1.48e+04}$ & $\num{9.98e-03}$ \\
& \ours & $\num{3.45e-06}\pm\num{8.18e-07}$ & $\num{7.50e-06}\pm\num{9.52e-06}$ & $\num{2.91e-08}\pm\num{8.68e-09}$ & $\num{1.85e+04}$ & $\num{1.22e-02}$ \\
\midrule
\multirow{2}{*}{9}
& GNS & $\num{5.67e-06}\pm\num{1.08e-06}$ & $\num{3.07e-05}\pm\num{3.60e-05}$ & $\num{1.82e-07}\pm\num{5.56e-08}$ & $\num{1.58e+04}$ & $\num{1.08e-02}$ \\
& \ours & $\num{2.95e-06}\pm\num{5.28e-07}$ & $\num{1.47e-05}\pm\num{1.61e-05}$ & $\num{7.71e-08}\pm\num{2.00e-08}$ & $\num{1.89e+04}$ & $\num{1.24e-02}$ \\
  \bottomrule
\end{tabular}
\end{table}

Based on the data of Table~\ref{tab:abl-hist}, we actually see that including history hurts performance. We expect this to be the case, since we do not expect second order position derivatives to be very physically meaningful at our timescale. As these do not provide a strong signal, it has the potential to confuse a model. Although the models perform best with a history of 2 (we do not test a history of 1, as this destroys velocity information), we still report results using a history length of 6 in all of our other experiments, as 6 is the LagrangeBench default, and truncation of the history fundamentally changes the prediction task. We attempt to align with the LagrangeBench default settings so our results are easily compared to others using LagrangeBench.

\subsubsection{Interpolation}
\label{app:intpl}
Let $D\in\mathbb{N}$ represent the spatial dimension. Then, let us denote the cell size $\Delta_d\in\mathbb{R}$ for each axis $d\in\llbracket 1, N \rrbracket$. Now, suppose our grid is confined by coordinate bounds $\mathbf{b}_{\min{}}, \mathbf{b}_{\max{}} \in \mathbb{R}^D$. Let the feature vector of a particle $p$ be $\mathbf{h}_p \in \mathbb{R}^H$ whose coordinates we denote $\mathbf{i}_p$. We use $\mathbf{G}_\mathbf{c} \in \mathbb{R}^H$ to represent the value at the grid cell whose index is the vector $\mathbf{c}\in\mathbb{N}^D$.

For brevity, let us define a normalization function operating on some coordinates $$\mathbf{u}(\mathbf{i}) = \frac{\mathbf{i} - \mathbf{b}_{\min{}}}{\mathbf{\Delta}}$$

Then, for each interpolation scheme whose 1-D kernel is denoted by $w(r)$, we use the multidimensional kernel $$K(\mathbf{i}, \mathbf{c}) = \prod_{d=1}^D w\left(\mathbf{u}(\mathbf{i})_d - \mathbf{c}_d - \frac{1}{2}\right)$$

Our scattering operation is thus defined as $$\mathbf{G}_\mathbf{c} = \sum_p K(\mathbf{i}_p, \mathbf{c})\mathbf{h}_p$$ and our gathering operation is defined as $$\mathbf{h}_p = \sum_\mathbf{c} K(\mathbf{i}_p, \mathbf{c}) \mathbf{G}_\mathbf{c}$$

For Nearest Grid Point \citep{evans1957pic}, we use the kernel $$w(r) = \begin{cases}
1,& |r|<\tfrac12,\\
0,& \text{otherwise.}
\end{cases}$$

For Cloud in Cell \citep{birdsall1969clouds}, we use the kernel $$w(r) = \max(0,\,1-|r|)$$

For Triangular Shaped Cloud \citep{eastwood1974shaping}, we use the kernel $$w(r) = \begin{cases}
0.75-r^2,& |r|<0.5,\\
0.5\,(1.5-|r|)^2,& 0.5\le |r|<1.5,\\
0,& |r|\ge 1.5.
\end{cases}$$

In Table~\ref{tab:abl-intpl}, we find that cloud in cell interpolation works well compared to other options. Surprisingly, using a TSC scatter operation with a NGP gather operation also works well, but we leave investigating this phenomenon further to future work.

\begin{table}[H]
\centering
\caption{Performance based on interpolation scheme}
\label{tab:abl-intpl}
\begin{tabular}{ccccccc}
  \toprule
  \textbf{Scatter} & \textbf{Gather} &
    \multicolumn{1}{c}{$\text{MSE}_{20}$} &
    \multicolumn{1}{c}{$\text{MSE}_{E_{\text{kin}}}$} &
    \multicolumn{1}{c}{Sinkhorn} \\
\midrule
NGP & NGP & $\num{1.90e-06}\pm\num{3.83e-07}$ & $\num{7.18e-06}\pm\num{7.76e-06}$ & $\num{6.40e-08}\pm\num{1.96e-08}$ \\
\midrule
NGP & CIC & $\num{1.89e-06}\pm\num{4.02e-07}$ & $\num{6.06e-06}\pm\num{6.89e-06}$ & $\num{5.31e-08}\pm\num{1.40e-08}$ \\
\midrule
NGP & TSC & $\num{1.79e-06}\pm\num{3.63e-07}$ & $\num{9.80e-06}\pm\num{1.19e-05}$ & $\num{6.44e-08}\pm\num{2.01e-08}$ \\
\midrule
CIC & NGP & $\num{2.10e-06}\pm\num{4.60e-07}$ & $\num{5.23e-06}\pm\num{5.48e-06}$ & $\num{4.64e-08}\pm\num{1.40e-08}$ \\
\midrule
CIC & CIC & $\num{1.56e-06}\pm\num{4.27e-07}$ & $\num{4.64e-06}\pm\num{5.05e-06}$ & $\num{4.45e-08}\pm\num{1.51e-08}$ \\
\midrule
CIC & TSC & $\num{1.81e-06}\pm\num{3.72e-07}$ & $\num{6.52e-06}\pm\num{7.03e-06}$ & $\num{6.15e-08}\pm\num{2.20e-08}$ \\
\midrule
TSC & NGP & $\num{1.55e-06}\pm\num{4.29e-07}$ & $\num{6.19e-06}\pm\num{7.44e-06}$ & $\num{2.87e-08}\pm\num{9.95e-09}$ \\
\midrule
TSC & CIC & $\num{1.90e-06}\pm\num{3.56e-07}$ & $\num{7.44e-06}\pm\num{8.87e-06}$ & $\num{5.27e-08}\pm\num{1.55e-08}$ \\
\midrule
TSC & TSC & $\num{1.74e-06}\pm\num{3.49e-07}$ & $\num{1.09e-05}\pm\num{1.17e-05}$ & $\num{6.13e-08}\pm\num{1.86e-08}$ \\
  \bottomrule
\end{tabular}
\end{table}

\subsection{Tuning}
After establishing the empirical soundness of our architecture, we further conduct experiments to probe the effect of various hyperparameters.

\subsubsection{Graph Connectivity}
We tested various settings for the graph connectivity radius. Although this is a parameter independent of the model itself, in the sense that this is done more as a data processing step rather than something neural in nature, it significantly affects the performance of our models. We present our findings in terms of the average node degree rather than as the raw connectivity radius so the results may be more interpretable between datasets; the connectivity radius determined in LagrangeBench is derived from the average distance to a neighbor, which is correlated to the average degree inversely.

\begin{figure}[H]
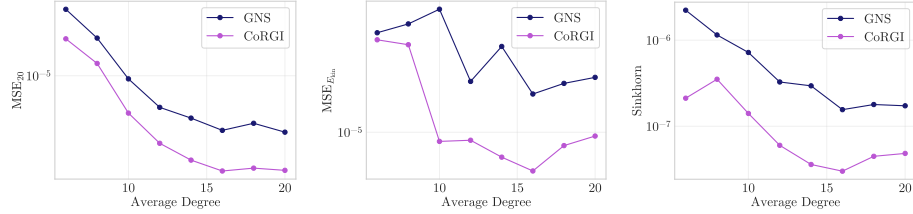

% \vspace{-0.1}
  \centering                  % centre the whole block
  % ----------- First panel -----------
  \begin{subfigure}[t]{0.32\textwidth} % 3 × 0.32 < 1 ⇒ room for spacing
    \includesvg[width=\linewidth]{figures/mse_20_vs_degree.svg}
    % \caption{$\text{MSE}_{20}$}
    % \label{fig:panelA}
  \end{subfigure}
  \hfill                         % stretchable horizontal space
  % ----------- Second panel ----------
  \begin{subfigure}[t]{0.32\textwidth}
    \includesvg[width=\linewidth]{figures/mse_ekin_vs_degree.svg}
    % \caption{Kinetic energy MSE}
    % \label{fig:panelB}
  \end{subfigure}
  \hfill
  % ----------- Third panel -----------
  \begin{subfigure}[t]{0.32\textwidth}
    \includesvg[width=\linewidth]{figures/sinkhorn_vs_degree.svg}
    % \caption{Sinkhorn}
    % \label{fig:panelC}
  \end{subfigure}

  \caption{Plot of accuracy compared to average node degree for GNS and \ours on RPF-2D.}
  % \label{fig:pareto}
\end{figure}

\begin{table}[H]
\centering
\caption{Performance based on average vertex degree}
\label{tab:abl-deg}
\begin{tabular}{ccccccc}
  \toprule
  \textbf{Degree} & \textbf{Model} &
    \multicolumn{1}{c}{$\text{MSE}_{20}$} &
    \multicolumn{1}{c}{$\text{MSE}_{E_{\text{kin}}}$} &
    \multicolumn{1}{c}{Sinkhorn} &
    \multicolumn{1}{c}{Training time} &
    \multicolumn{1}{c}{Inference time} \\
\midrule
\multirow{2}{*}{6}
& GNS & $\num{3.64e-05}\pm\num{4.03e-06}$ & $\num{4.89e-05}\pm\num{5.41e-05}$ & $\num{2.23e-06}\pm\num{6.68e-07}$ & $\num{1.43e+04}$ & $\num{1.06e-02}$ \\
& \ours & $\num{2.05e-05}\pm\num{1.76e-06}$ & $\num{4.37e-05}\pm\num{4.34e-05}$ & $\num{2.12e-07}\pm\num{5.36e-08}$ & $\num{1.90e+04}$ & $\num{1.24e-02}$ \\
\midrule
\multirow{2}{*}{8}
& GNS & $\num{2.08e-05}\pm\num{2.81e-06}$ & $\num{5.63e-05}\pm\num{6.35e-05}$ & $\num{1.15e-06}\pm\num{2.88e-07}$ & $\num{1.38e+04}$ & $\num{9.78e-03}$ \\
& \ours & $\num{1.27e-05}\pm\num{1.41e-06}$ & $\num{4.04e-05}\pm\num{3.84e-05}$ & $\num{3.52e-07}\pm\num{9.01e-08}$ & $\num{1.81e+04}$ & $\num{1.17e-02}$ \\
\midrule
\multirow{2}{*}{10}
& GNS & $\num{9.41e-06}\pm\num{1.81e-06}$ & $\num{7.12e-05}\pm\num{7.17e-05}$ & $\num{7.20e-07}\pm\num{1.90e-07}$ & $\num{1.55e+04}$ & $\num{1.10e-02}$ \\
& \ours & $\num{4.84e-06}\pm\num{8.09e-07}$ & $\num{8.64e-06}\pm\num{9.91e-06}$ & $\num{1.41e-07}\pm\num{3.97e-08}$ & $\num{1.89e+04}$ & $\num{1.22e-02}$ \\
\midrule
\multirow{2}{*}{12}
& GNS & $\num{5.41e-06}\pm\num{1.07e-06}$ & $\num{2.25e-05}\pm\num{2.44e-05}$ & $\num{3.27e-07}\pm\num{1.15e-07}$ & $\num{1.50e+04}$ & $\num{1.06e-02}$ \\
& \ours & $\num{2.69e-06}\pm\num{5.78e-07}$ & $\num{8.80e-06}\pm\num{9.05e-06}$ & $\num{6.01e-08}\pm\num{1.74e-08}$ & $\num{1.72e+04}$ & $\num{1.11e-02}$ \\
\midrule
\multirow{2}{*}{14}
& GNS & $\num{4.39e-06}\pm\num{8.48e-07}$ & $\num{3.93e-05}\pm\num{4.29e-05}$ & $\num{2.95e-07}\pm\num{1.10e-07}$ & $\num{1.54e+04}$ & $\num{1.05e-02}$ \\
& \ours & $\num{1.94e-06}\pm\num{4.45e-07}$ & $\num{6.72e-06}\pm\num{7.36e-06}$ & $\num{3.60e-08}\pm\num{1.05e-08}$ & $\num{1.83e+04}$ & $\num{1.21e-02}$ \\
\midrule
\multirow{2}{*}{16}
& GNS & $\num{3.46e-06}\pm\num{7.44e-07}$ & $\num{1.84e-05}\pm\num{1.98e-05}$ & $\num{1.56e-07}\pm\num{6.04e-08}$ & $\num{1.51e+04}$ & $\num{1.00e-02}$ \\
& \ours & $\num{1.57e-06}\pm\num{4.18e-07}$ & $\num{5.39e-06}\pm\num{6.10e-06}$ & $\num{3.01e-08}\pm\num{1.03e-08}$ & $\num{1.79e+04}$ & $\num{1.15e-02}$ \\
\midrule
\multirow{2}{*}{18}
& GNS & $\num{3.97e-06}\pm\num{7.34e-07}$ & $\num{2.18e-05}\pm\num{2.37e-05}$ & $\num{1.79e-07}\pm\num{6.91e-08}$ & $\num{1.57e+04}$ & $\num{1.01e-02}$ \\
& \ours & $\num{1.66e-06}\pm\num{3.41e-07}$ & $\num{8.08e-06}\pm\num{9.15e-06}$ & $\num{4.48e-08}\pm\num{1.50e-08}$ & $\num{1.76e+04}$ & $\num{1.10e-02}$ \\
\midrule
\multirow{2}{*}{20}
& GNS & $\num{3.34e-06}\pm\num{7.09e-07}$ & $\num{2.40e-05}\pm\num{2.75e-05}$ & $\num{1.73e-07}\pm\num{7.74e-08}$ & $\num{1.59e+04}$ & $\num{1.07e-02}$ \\
& \ours & $\num{1.59e-06}\pm\num{3.09e-07}$ & $\num{9.41e-06}\pm\num{1.07e-05}$ & $\num{4.84e-08}\pm\num{1.49e-08}$ & $\num{1.86e+04}$ & $\num{1.20e-02}$ \\
  \bottomrule
\end{tabular}
\end{table}

We see that \ours consistently outperforms GNS regardless of the average node degree of the input graph. We also see that the performance seems to peak when the graph connectivity is approximately 16, which is similar to the default provided in LagrangeBench. We expect to see that there is no significant difference in training and inference time between the different connectivity settings due to the edge parallelization of GNS, which Table~\ref{tab:abl-deg} empirically demonstrates. We additionally note that the performance benefit of \ours persists regardless of the connectivity of the graph.

\subsubsection{Convolution Parameters}
In order to choose the parameters for our global convolution module, we conduct a sweep through various depths and base widths. By depth, we mean the number of different convolutional resolutions involved. For instance, the depiction in Fig~\ref{fig:arch} has a depth of 3. The test with a depth of 0 serves as an ablation, verifying that our improvements indeed come from genuine convolution operations rather than simply projecting to and from a grid. By width, we refer to the number of channels in the highest resolution layer. At each layer, we double the width and concatenate a skip connection from the graph output of the encoder module.

\begin{table}[H]
\centering
\caption{Performance based on convolution depth and width.}
\label{tab:abl-conv}
  \begin{tabular}{ccccccc}
  % --------------------------------

    \toprule
    \textbf{Depth} & \textbf{Width} &
      \multicolumn{1}{c}{$\text{MSE}_{20}$ ($\times 10^{-6}$)} &
      \multicolumn{1}{c}{$\text{MSE}_{E_{\text{kin}}}$} &
      \multicolumn{1}{c}{Sinkhorn} &
      \multicolumn{1}{c}{Training time (s)} &
      \multicolumn{1}{c}{Inference time (s)} \\
    \midrule
    % depth = 0 (width irrelevant)
    0 & \multicolumn{1}{c}{--} & {$\num{3.91e-06}\pm\num{9.73e-07}$} & {$\num{3.29e-05}\pm\num{4.74e-05}$} & {$\num{1.46e-07}\pm\num{7.49e-8}$} & {\num{9.72e-03}} & \num{1.49e+04} \\
\midrule
\multirow{7}{*}{1}
& 32 & $\num{3.18e-06}\pm\num{7.35e-07}$ & $\num{1.23e-05}\pm\num{1.52e-05}$ & $\num{1.61e-07}\pm\num{6.52e-08}$ & $\num{1.58e+04}$ & $\num{1.08e-02}$ \\
& 48 & $\num{3.19e-06}\pm\num{7.21e-07}$ & $\num{1.23e-05}\pm\num{1.47e-05}$ & $\num{1.83e-07}\pm\num{7.43e-08}$ & $\num{1.57e+04}$ & $\num{1.08e-02}$ \\
& 64 & $\num{3.06e-06}\pm\num{6.79e-07}$ & $\num{9.87e-06}\pm\num{1.06e-05}$ & $\num{1.53e-07}\pm\num{6.76e-08}$ & $\num{1.47e+04}$ & $\num{9.89e-03}$ \\
& 96 & $\num{2.96e-06}\pm\num{6.82e-07}$ & $\num{9.22e-06}\pm\num{1.05e-05}$ & $\num{1.57e-07}\pm\num{6.08e-08}$ & $\num{1.56e+04}$ & $\num{1.03e-02}$ \\
& 128 & $\num{2.91e-06}\pm\num{6.46e-07}$ & $\num{9.79e-06}\pm\num{1.07e-05}$ & $\num{1.78e-07}\pm\num{8.26e-08}$ & $\num{1.65e+04}$ & $\num{1.10e-02}$ \\
& 192 & $\num{3.03e-06}\pm\num{6.45e-07}$ & $\num{2.10e-05}\pm\num{2.10e-05}$ & $\num{2.11e-07}\pm\num{1.12e-07}$ & $\num{1.63e+04}$ & $\num{1.12e-02}$ \\
& 256 & $\num{3.14e-06}\pm\num{7.02e-07}$ & $\num{1.10e-05}\pm\num{1.20e-05}$ & $\num{1.71e-07}\pm\num{7.61e-08}$ & $\num{1.62e+04}$ & $\num{1.07e-02}$ \\
\midrule
\multirow{7}{*}{2}
& 32 & $\num{2.85e-06}\pm\num{6.39e-07}$ & $\num{1.67e-05}\pm\num{1.84e-05}$ & $\num{1.49e-07}\pm\num{6.51e-08}$ & $\num{1.56e+04}$ & $\num{1.04e-02}$ \\
& 48 & $\num{3.15e-06}\pm\num{5.97e-07}$ & $\num{1.32e-05}\pm\num{1.41e-05}$ & $\num{1.58e-07}\pm\num{5.26e-08}$ & $\num{1.74e+04}$ & $\num{1.16e-02}$ \\
& 64 & $\num{2.68e-06}\pm\num{5.87e-07}$ & $\num{1.34e-05}\pm\num{1.49e-05}$ & $\num{1.24e-07}\pm\num{5.59e-08}$ & $\num{1.65e+04}$ & $\num{1.11e-02}$ \\
& 96 & $\num{2.54e-06}\pm\num{5.73e-07}$ & $\num{1.02e-05}\pm\num{1.17e-05}$ & $\num{9.39e-08}\pm\num{3.15e-08}$ & $\num{1.71e+04}$ & $\num{1.17e-02}$ \\
& 128 & $\num{2.92e-06}\pm\num{6.01e-07}$ & $\num{1.06e-05}\pm\num{1.15e-05}$ & $\num{1.23e-07}\pm\num{4.57e-08}$ & $\num{1.72e+04}$ & $\num{1.18e-02}$ \\
& 192 & $\num{2.40e-06}\pm\num{5.30e-07}$ & $\num{9.60e-06}\pm\num{1.14e-05}$ & $\num{9.24e-08}\pm\num{3.43e-08}$ & $\num{1.65e+04}$ & $\num{1.06e-02}$ \\
& 256 & $\num{2.41e-06}\pm\num{5.79e-07}$ & $\num{9.76e-06}\pm\num{1.05e-05}$ & $\num{9.71e-08}\pm\num{3.46e-08}$ & $\num{1.78e+04}$ & $\num{1.07e-02}$ \\
\midrule
\multirow{7}{*}{3}
& 32 & $\num{1.69e-06}\pm\num{4.38e-07}$ & $\num{6.97e-06}\pm\num{7.61e-06}$ & $\num{3.87e-08}\pm\num{1.54e-08}$ & $\num{1.63e+04}$ & $\num{1.08e-02}$ \\
& 48 & $\num{1.75e-06}\pm\num{4.69e-07}$ & $\num{6.54e-06}\pm\num{7.00e-06}$ & $\num{4.36e-08}\pm\num{1.71e-08}$ & $\num{1.67e+04}$ & $\num{1.09e-02}$ \\
& 64 & $\num{1.64e-06}\pm\num{3.50e-07}$ & $\num{6.91e-06}\pm\num{8.01e-06}$ & $\num{4.52e-08}\pm\num{1.32e-08}$ & $\num{1.78e+04}$ & $\num{1.16e-02}$ \\
& 96 & $\num{1.56e-06}\pm\num{4.32e-07}$ & $\num{7.14e-06}\pm\num{8.31e-06}$ & $\num{2.99e-08}\pm\num{1.11e-08}$ & $\num{1.65e+04}$ & $\num{1.07e-02}$ \\
& 128 & $\num{1.53e-06}\pm\num{4.14e-07}$ & $\num{4.61e-06}\pm\num{4.80e-06}$ & $\num{2.51e-08}\pm\num{7.91e-09}$ & $\num{1.75e+04}$ & $\num{1.11e-02}$ \\
& 192 & $\num{1.77e-06}\pm\num{3.77e-07}$ & $\num{8.21e-06}\pm\num{9.21e-06}$ & $\num{8.13e-08}\pm\num{1.97e-08}$ & $\num{1.94e+04}$ & $\num{1.18e-02}$ \\
& 256 & $\num{1.69e-06}\pm\num{3.50e-07}$ & $\num{6.14e-06}\pm\num{7.45e-06}$ & $\num{4.49e-08}\pm\num{1.33e-08}$ & $\num{2.15e+04}$ & $\num{1.20e-02}$ \\
\midrule
\multirow{7}{*}{4}
& 32 & $\num{1.69e-06}\pm\num{4.26e-07}$ & $\num{6.07e-06}\pm\num{6.75e-06}$ & $\num{3.07e-08}\pm\num{7.75e-09}$ & $\num{1.88e+04}$ & $\num{1.26e-02}$ \\
& 48 & $\num{1.59e-06}\pm\num{4.09e-07}$ & $\num{4.92e-06}\pm\num{5.86e-06}$ & $\num{2.28e-08}\pm\num{5.88e-09}$ & $\num{1.89e+04}$ & $\num{1.20e-02}$ \\
& 64 & $\num{1.88e-06}\pm\num{3.87e-07}$ & $\num{7.42e-06}\pm\num{8.50e-06}$ & $\num{4.53e-08}\pm\num{1.02e-08}$ & $\num{1.90e+04}$ & $\num{1.25e-02}$ \\
& 96 & $\num{1.77e-06}\pm\num{3.58e-07}$ & $\num{1.22e-05}\pm\num{1.40e-05}$ & $\num{3.60e-08}\pm\num{8.79e-09}$ & $\num{1.95e+04}$ & $\num{1.25e-02}$ \\
& 128 & $\num{1.56e-06}\pm\num{4.08e-07}$ & $\num{4.17e-06}\pm\num{4.37e-06}$ & $\num{2.20e-08}\pm\num{5.40e-09}$ & $\num{2.06e+04}$ & $\num{1.20e-02}$ \\
& 192 & $\num{1.52e-06}\pm\num{4.02e-07}$ & $\num{4.31e-06}\pm\num{4.83e-06}$ & $\num{1.72e-08}\pm\num{4.94e-09}$ & $\num{2.52e+04}$ & $\num{1.25e-02}$ \\
& 256 & $\num{1.67e-06}\pm\num{3.55e-07}$ & $\num{5.89e-06}\pm\num{6.66e-06}$ & $\num{3.70e-08}\pm\num{1.03e-08}$ & $\num{3.10e+04}$ & $\num{1.32e-02}$ \\
    \bottomrule
  \end{tabular}
\end{table}

Matching our analysis in App~\ref{app: theory}, we find that a depth of 3 or 4 is optimal, as we obtain a roughly global receptive field given our initial resolution on RPF-2D of $32 \times 64$.

We moreover show that increasing our width has diminishing returns, and in particular, beyond a width of 128 on depths of 3 and 4 we actually see a degradation in performance. Since we utilized a backbone of GNS with a latent dimension of 128, we believe that the choice of width in our global convolution module should simply be inherited from the graph encoder module.

We note that increasing the width approximately maintains the inference time, but generally increases the training time. Predictably, increasing the depth will increase both the inference time and the training time.

We also hypothesize that the density of our grid has a significant impact on performance. We attempt to justify empirically that our choice (i.e. a density of one or two particles per voxel) in Table~\ref{tab:abl-res}.

\begin{table}[H]
\centering
\caption{Performance based on convolution resolution}
\label{tab:abl-res}
\begin{tabular}{ccccccc}
  \toprule
  \textbf{Resolution} &
    \multicolumn{1}{c}{$\text{MSE}_{20}$} &
    \multicolumn{1}{c}{$\text{MSE}_{E_{\text{kin}}}$} &
    \multicolumn{1}{c}{Sinkhorn} &
    \multicolumn{1}{c}{Training time} &
    \multicolumn{1}{c}{Inference time} \\
  \midrule
$16\times32$ & $\num{1.78e-06}\pm\num{3.58e-07}$ & $\num{6.17e-06}\pm\num{7.23e-06}$ & $\num{3.43e-08}\pm\num{8.05e-09}$ & $\num{1.72e+04}$ & $\num{1.08e-02}$ \\
\midrule
$32\times64$ & $\num{1.54e-06}\pm\num{4.17e-07}$ & $\num{4.58e-06}\pm\num{5.33e-06}$ & $\num{2.37e-08}\pm\num{7.71e-09}$ & $\num{1.72e+04}$ & $\num{1.09e-02}$ \\
\midrule
$48\times96$ & $\num{2.57e-06}\pm\num{4.71e-07}$ & $\num{1.25e-05}\pm\num{1.43e-05}$ & $\num{1.59e-07}\pm\num{5.60e-08}$ & $\num{1.78e+04}$ & $\num{1.04e-02}$ \\
\midrule
$64\times128$ & $\num{2.49e-06}\pm\num{5.77e-07}$ & $\num{1.33e-05}\pm\num{1.53e-05}$ & $\num{1.09e-07}\pm\num{4.69e-08}$ & $\num{1.94e+04}$ & $\num{1.07e-02}$ \\
  \bottomrule
\end{tabular}
\end{table}

The data show that our choice of $32\times 64$ performs well given our other default settings and our chosen aspect ratio. The drastic decrease in performance in higher resolutions compared to the relatively minor decrease in performance in the lower resolution corroborates our claim that we need for nodes to have a global receptive field: higher resolutions require more than 3 layers of global convolution, but in Table~\ref{tab:abl-res}, we test using only 3 layers.

\subsubsection{Learning Parameters}
We further assess the impact of \ours on learning parameters.

\begin{table}[H]
\centering
\caption{Performance based on learning rate and batch size}
\label{tab:abl-lr}
  \begin{tabular}{ccccccc}
  % --------------------------------

    \toprule
    \textbf{Batch Size} & \textbf{Learning Rate} &
      \multicolumn{1}{c}{$\text{MSE}_{20}$} &
      \multicolumn{1}{c}{$\text{MSE}_{E_{\text{kin}}}$} &
      \multicolumn{1}{c}{Sinkhorn} &
      \multicolumn{1}{c}{Training time (s)} &
      \multicolumn{1}{c}{Inference time (s)} \\
\midrule
\multirow{4}{*}{1}
& \num{5e-05} & $\num{4.63e-06}\pm\num{7.47e-07}$ & $\num{9.03e-06}\pm\num{9.79e-06}$ & $\num{1.57e-07}\pm\num{4.28e-08}$ & $\num{1.83e+04}$ & $\num{1.21e-02}$ \\
& \num{1e-04} & $\num{2.55e-06}\pm\num{5.28e-07}$ & $\num{8.05e-06}\pm\num{9.42e-06}$ & $\num{6.81e-08}\pm\num{2.06e-08}$ & $\num{1.89e+04}$ & $\num{1.23e-02}$ \\
& \num{5e-04} & $\num{1.75e-06}\pm\num{3.49e-07}$ & $\num{7.54e-06}\pm\num{8.93e-06}$ & $\num{4.87e-08}\pm\num{1.38e-08}$ & $\num{1.87e+04}$ & $\num{1.22e-02}$ \\
& \num{1e-03} & $\num{1.51e-06}\pm\num{3.13e-07}$ & $\num{4.91e-06}\pm\num{5.22e-06}$ & $\num{3.60e-08}\pm\num{9.26e-09}$ & $\num{1.76e+04}$ & $\num{1.09e-02}$ \\
\midrule
\multirow{4}{*}{2}
& \num{5e-05} & $\num{4.36e-06}\pm\num{7.24e-07}$ & $\num{1.31e-05}\pm\num{1.49e-05}$ & $\num{1.17e-07}\pm\num{3.06e-08}$ & $\num{1.78e+04}$ & $\num{1.14e-02}$ \\
& \num{1e-04} & $\num{2.52e-06}\pm\num{4.87e-07}$ & $\num{5.94e-06}\pm\num{5.55e-06}$ & $\num{6.08e-08}\pm\num{1.74e-08}$ & $\num{1.83e+04}$ & $\num{1.15e-02}$ \\
& \num{5e-04} & $\num{1.56e-06}\pm\num{4.00e-07}$ & $\num{6.63e-06}\pm\num{7.68e-06}$ & $\num{3.61e-08}\pm\num{1.36e-08}$ & $\num{1.77e+04}$ & $\num{1.12e-02}$ \\
& \num{1e-03} & $\num{1.53e-06}\pm\num{3.21e-07}$ & $\num{8.93e-06}\pm\num{1.04e-05}$ & $\num{3.53e-08}\pm\num{1.08e-08}$ & $\num{1.73e+04}$ & $\num{1.10e-02}$ \\
\midrule
\multirow{4}{*}{4}
& \num{5e-05} & $\num{4.48e-06}\pm\num{7.58e-07}$ & $\num{1.36e-05}\pm\num{1.37e-05}$ & $\num{1.00e-07}\pm\num{2.69e-08}$ & $\num{1.88e+04}$ & $\num{1.23e-02}$ \\
& \num{1e-04} & $\num{2.48e-06}\pm\num{4.93e-07}$ & $\num{6.00e-06}\pm\num{6.32e-06}$ & $\num{5.96e-08}\pm\num{1.68e-08}$ & $\num{1.73e+04}$ & $\num{1.10e-02}$ \\
& \num{5e-04} & $\num{1.54e-06}\pm\num{4.15e-07}$ & $\num{6.24e-06}\pm\num{7.34e-06}$ & $\num{2.35e-08}\pm\num{8.45e-09}$ & $\num{1.89e+04}$ & $\num{1.22e-02}$ \\
& \num{1e-03} & $\num{1.57e-06}\pm\num{3.19e-07}$ & $\num{5.33e-06}\pm\num{6.72e-06}$ & $\num{4.89e-08}\pm\num{1.46e-08}$ & $\num{1.72e+04}$ & $\num{1.04e-02}$ \\
\midrule
\multirow{4}{*}{8}
& \num{5e-05} & $\num{4.28e-06}\pm\num{6.91e-07}$ & $\num{1.11e-05}\pm\num{1.13e-05}$ & $\num{9.82e-08}\pm\num{2.42e-08}$ & $\num{1.85e+04}$ & $\num{1.22e-02}$ \\
& \num{1e-04} & $\num{2.49e-06}\pm\num{4.97e-07}$ & $\num{6.69e-06}\pm\num{7.58e-06}$ & $\num{7.26e-08}\pm\num{1.96e-08}$ & $\num{1.85e+04}$ & $\num{1.21e-02}$ \\
& \num{5e-04} & $\num{1.51e-06}\pm\num{3.91e-07}$ & $\num{6.02e-06}\pm\num{6.35e-06}$ & $\num{2.47e-08}\pm\num{8.76e-09}$ & $\num{1.88e+04}$ & $\num{1.24e-02}$ \\
& \num{1e-03} & $\num{1.56e-06}\pm\num{3.41e-07}$ & $\num{8.69e-06}\pm\num{1.07e-05}$ & $\num{4.79e-08}\pm\num{1.35e-08}$ & $\num{1.62e+04}$ & $\num{9.89e-03}$ \\
    \bottomrule
  \end{tabular}
\end{table}

In Table~\ref{tab:abl-lr}, we find empirically that our optimal learning rates match that of LagrangeBench. We expect that \ours will not require a significant change in learning parameters, reducing much experimental overhead in incorporating \ours in real world prediction tasks.

\subsection{Longer Times}
In this section, we qualitatively evaluate our performance over longer time frames in the DAM-2D dataset. We choose this dataset as its high turbulence represents a more realistic usecase and is more succeptible to the butterfly effect, hence making long term predictions more challenging.

We compare our performance to those found in the Neural SPH paper \citep{toshev2024neural}, but are unable to reproduce their results: despite multiple attempted revisions, we could not successfully run the code provided by Toshev et al. Their code includes references to modules not included in the camera ready version, and hence, is not executable in its released form. As we could not reproduce their experiments, we simply restate their results and compare them against our own. They did not report significance, hence it is not present in our table. We borrow their notation, where $\square_g$ has external forces removed and $\square_p$ includes a pressure term. 

\begin{table}[H]
\centering
\caption{Performance compared to Neural SPH.}
\label{tab:abl-sph}
  \begin{tabular}{ccccccc}
  % --------------------------------

    \toprule
    \textbf{Model} &
      \multicolumn{1}{c}{$\text{MSE}_{400}$} &
      \multicolumn{1}{c}{$\text{MSE}_{E_{\text{kin}}}$} &
      \multicolumn{1}{c}{Sinkhorn} \\
\midrule
$\text{GNS}$ & $\num{1.05e-1}\pm\num{2.05e-2}$ & $\num{2.22e-2}\pm\num{1.24e-02}$ & $\num{1.91e-2}\pm\num{9.57e-3}$ \\
$\text{GNS}_g$ & \num{8.0e-2} & \num{1.3e-2} & \num{9.4e-3} \\
$\text{GNS}_p$ & \num{9.7e-2} & \num{7.1e-3} & \num{5.8e-3} \\
$\text{GNS}_{g,p}$ & \num{8.4e-2} & \num{7.5e-3} & \num{2.1e-3} \\
\midrule
$\text{SEGNN}$ & $\num{1.71e-1}\pm\num{1.01e-2}$ & $\num{1.21e-2}\pm\num{1.83e-03}$ & $\num{3.09e-2}\pm\num{5.98e-3}$ \\
$\text{SEGNN}_g$ & \num{1.6e-1} & \num{2.1e-2} & \num{1.9e+1} \\
$\text{SEGNN}_p$ & \num{1.2e-1} & \num{9.4e-3} & \num{5.2e-2} \\
$\text{SEGNN}_{g,p}$ & \num{8.6e-2} & \num{4.9e-3} & \num{2.6e-3} \\
\midrule
\ours & $\num{2.84e-2}\pm\num{2.53e-3}$ & $\num{3.04e-4}\pm\num{8.79e-5}$ & $\num{9.59e-4}\pm\num{1.96e-4}$ \\
    \bottomrule
  \end{tabular}
\end{table}

From our experiments, we conclude that \ours performs substantially better on the DAM-2D dataset. However, it is possible to use Neural SPH with \ours in theory, and we can potentially see even better performance. We additionally provide qualitative analysis of how well \ours performs with longer rollouts.

\begin{figure}[H]
  \centering                  % centre the whole block
  % ----------- First panel -----------
  \begin{subfigure}[t]{0.24\textwidth} % 3 × 0.32 < 1 ⇒ room for spacing
    \includegraphics[width=\linewidth]{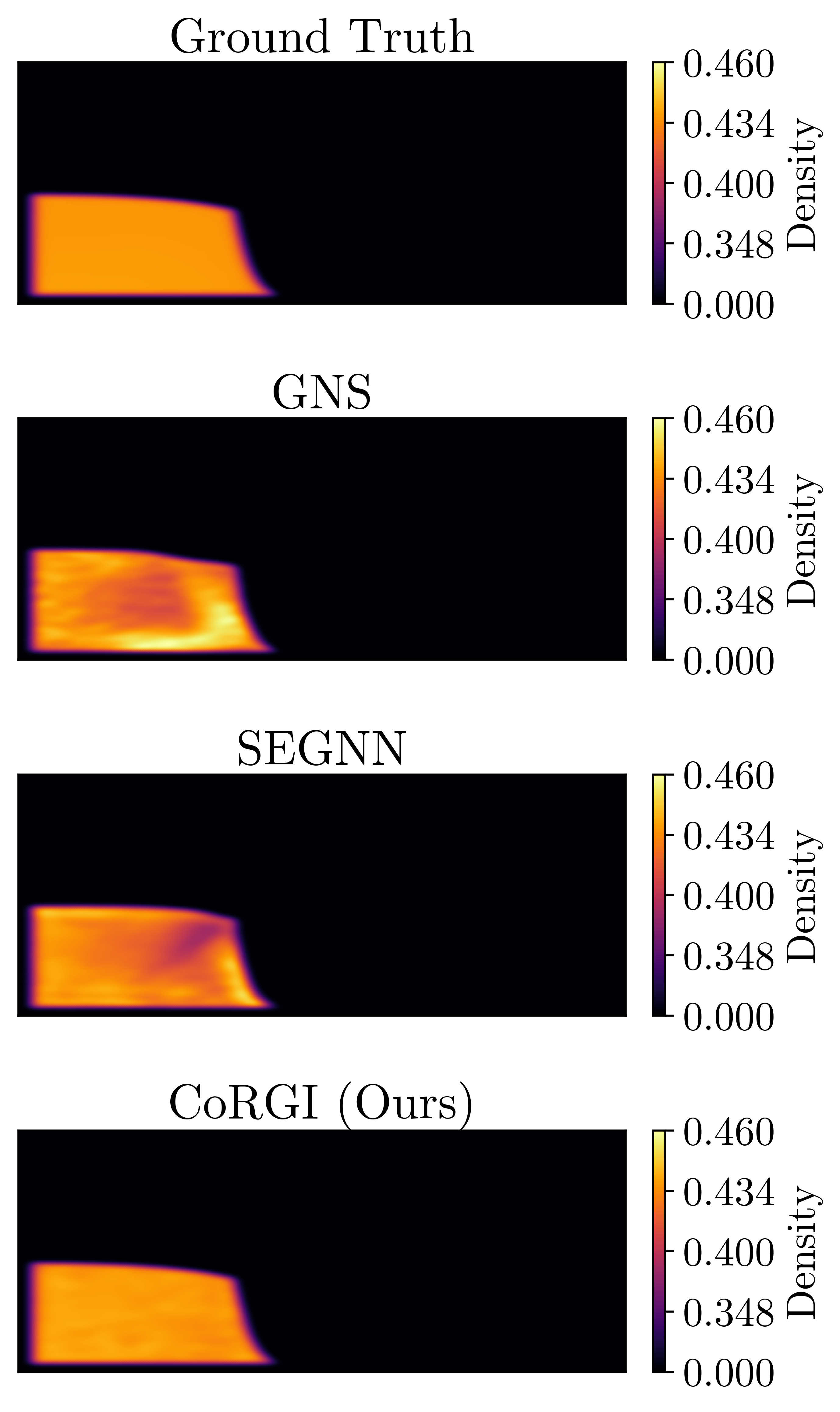}
    \caption{$t = 20$}
    % \label{fig:panelA}
  \end{subfigure}
  % \hfill
  \begin{subfigure}[t]{0.24\textwidth}
    \includegraphics[width=\linewidth]{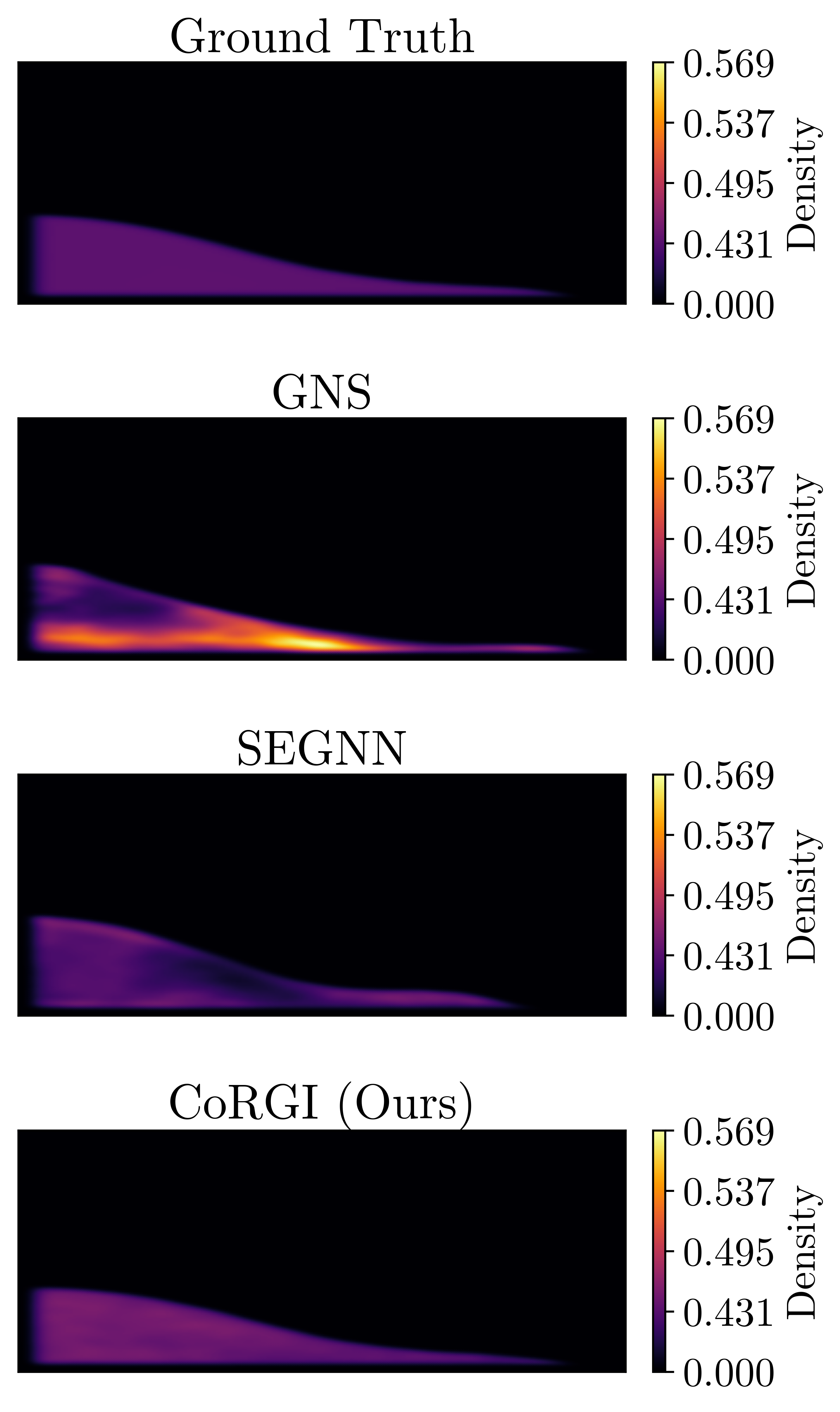}
    \caption{$t = 80$}
  \end{subfigure}
  % \hfill
  % ----------- Second panel ----------
  \begin{subfigure}[t]{0.24\textwidth}
    \includegraphics[width=\linewidth]{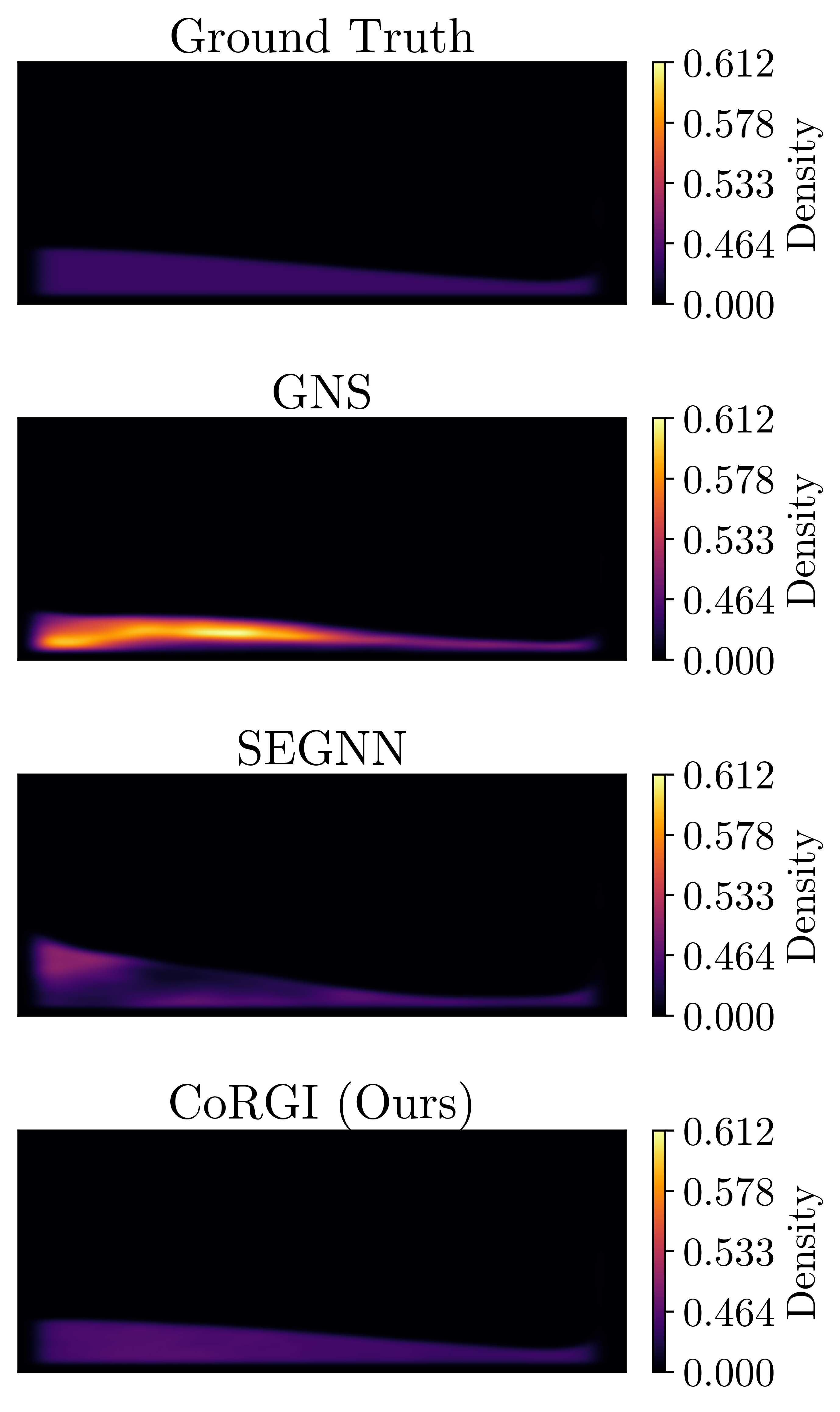}
    \caption{$t = 140$}
    % \label{fig:panelB}
  \end{subfigure}
  \begin{subfigure}[t]{0.24\textwidth} % 3 × 0.32 < 1 ⇒ room for spacing
    \includegraphics[width=\linewidth]{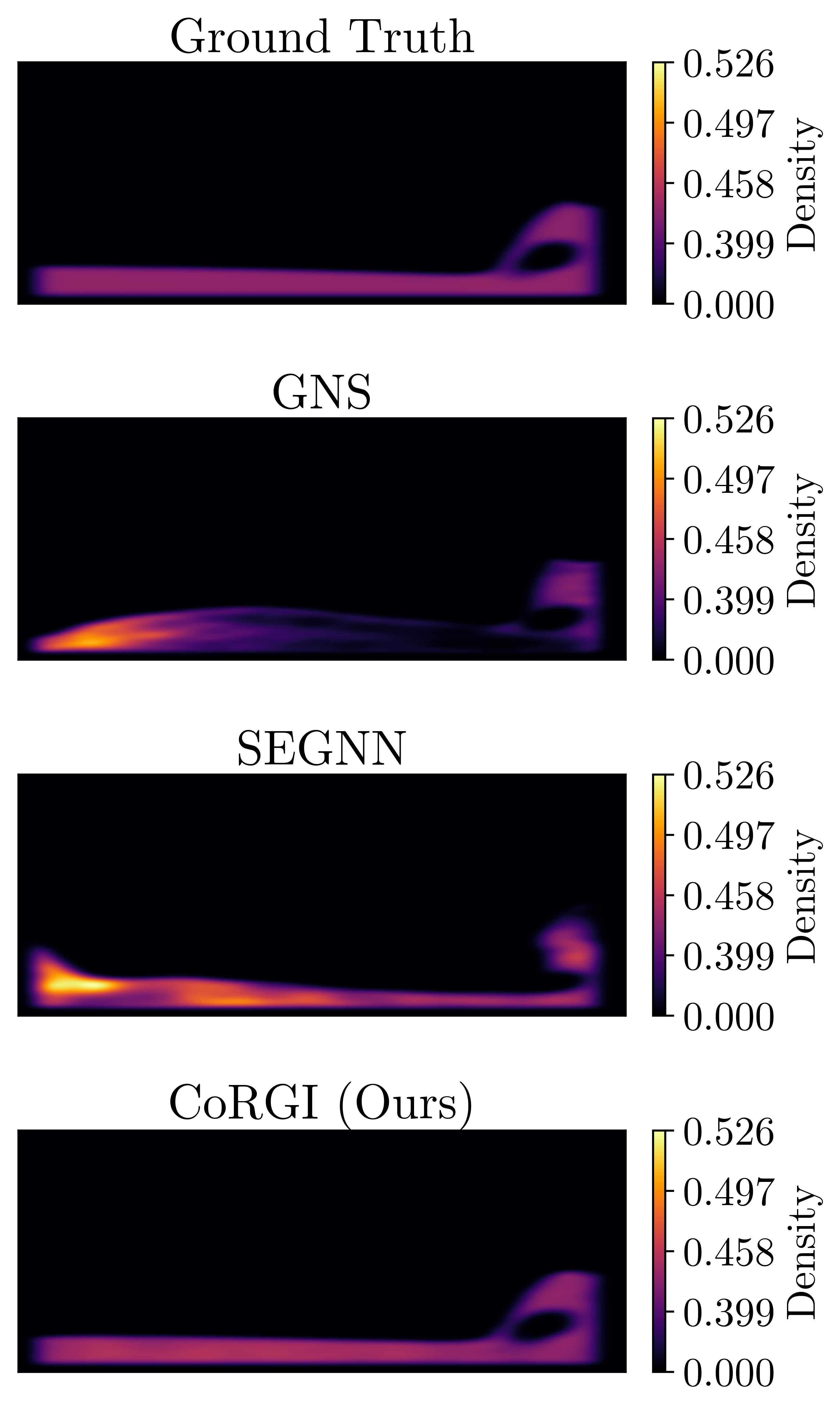}
    \caption{$t = 200$}
    % \label{fig:panelA}
  \end{subfigure}
  % \hfill
  \begin{subfigure}[t]{0.32\textwidth}
    \includegraphics[width=\linewidth]{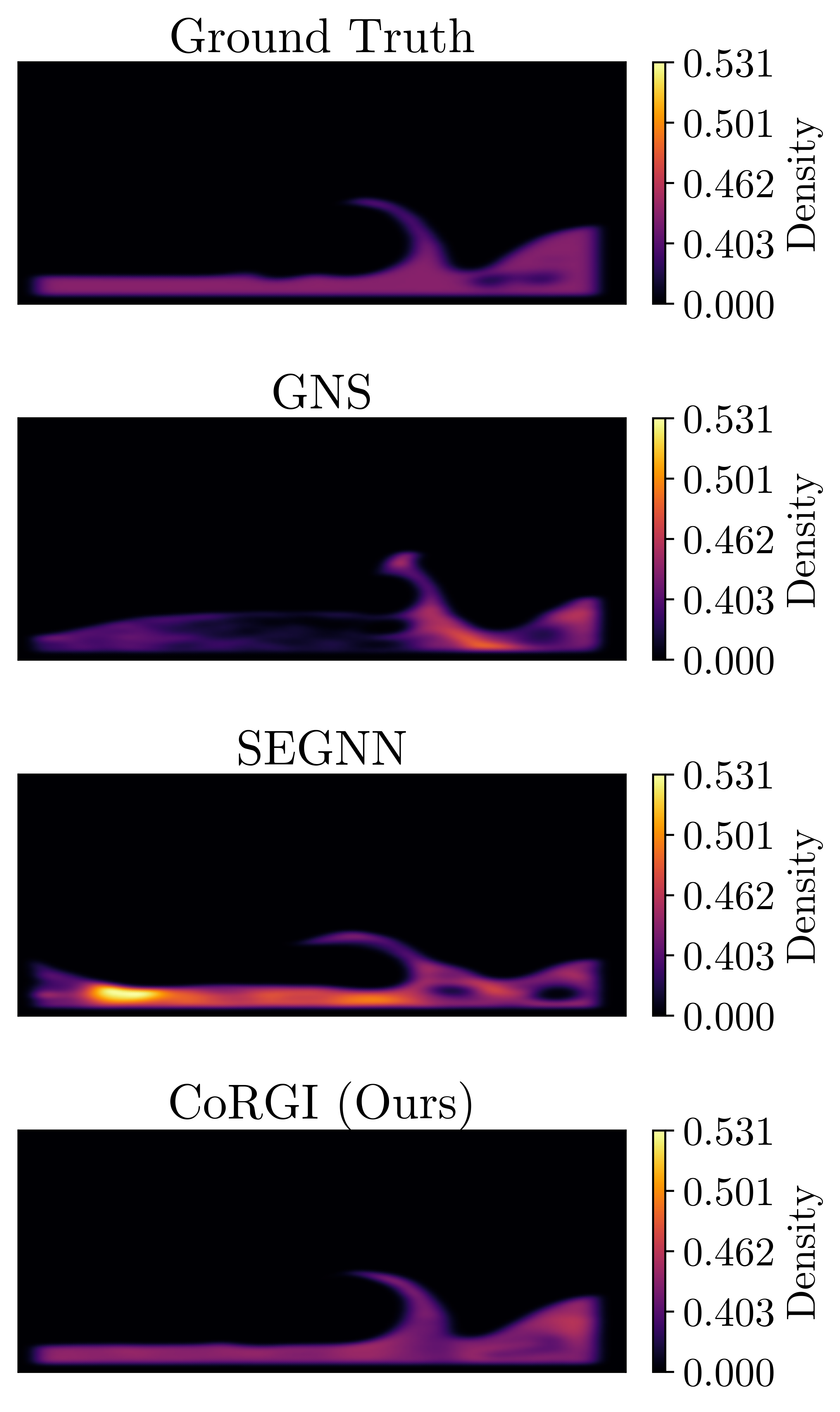}
    \caption{$t = 260$}
  \end{subfigure}
  \hfill
  % ----------- Second panel ----------
  \begin{subfigure}[t]{0.32\textwidth}
    \includegraphics[width=\linewidth]{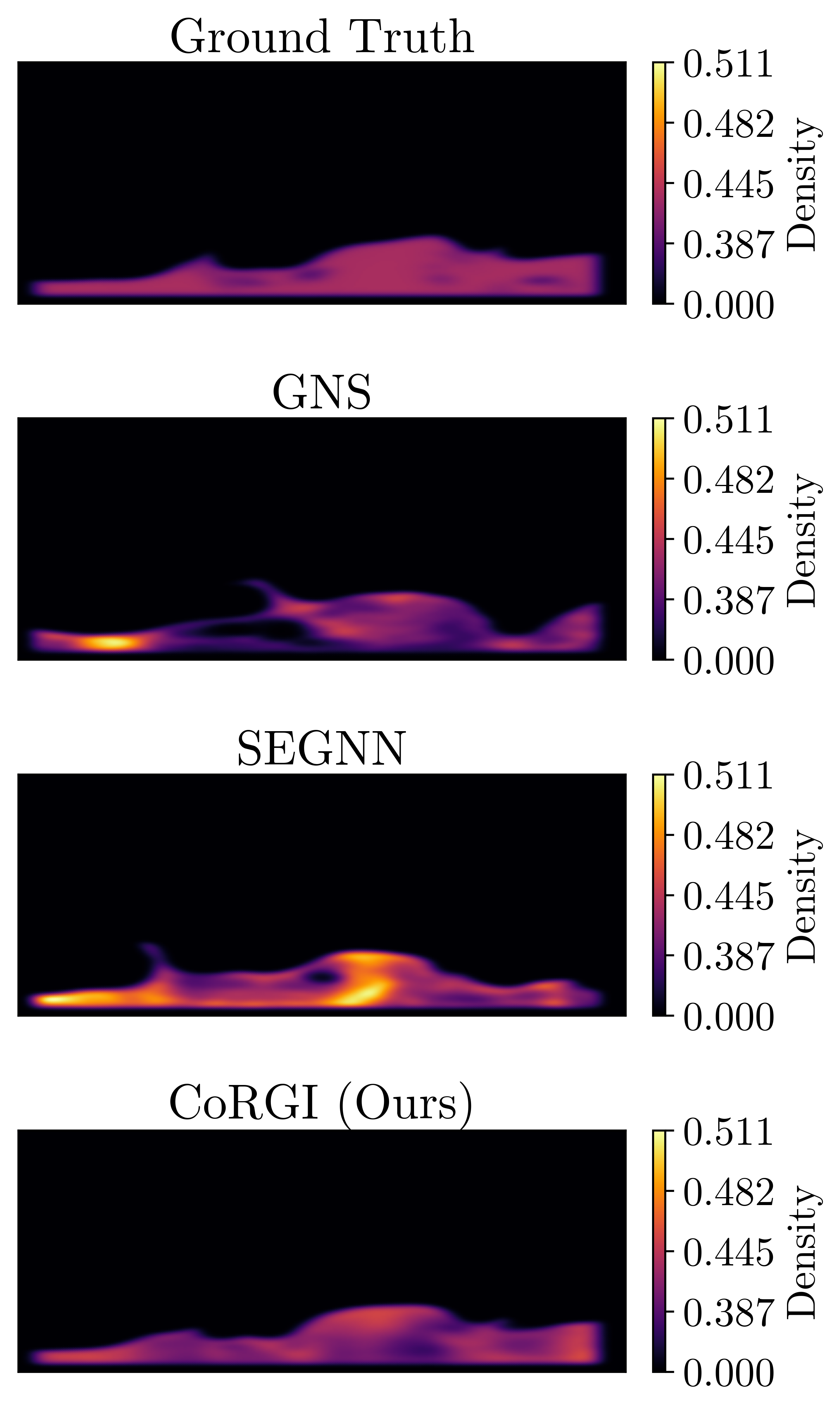}
    \caption{$t = 320$}
    % \label{fig:panelB}
  \end{subfigure}
  \hfill
    \begin{subfigure}[t]{0.32\textwidth}
    \includegraphics[width=\linewidth]{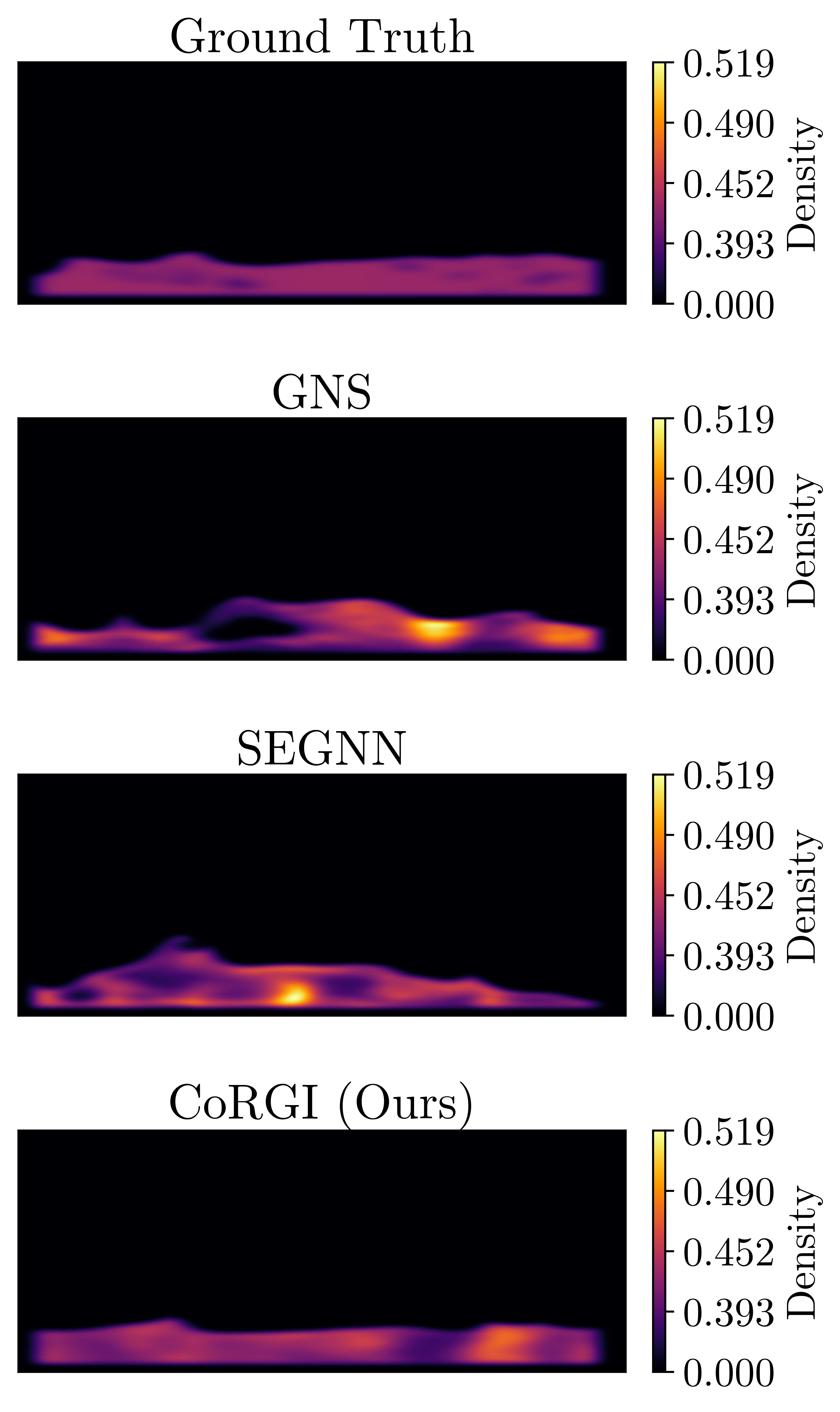}
    \caption{$t = 380$}
    % \label{fig:panelB}
  \end{subfigure}

  \caption{Kernel density estimation on DAM-2D ($t \in \llbracket 0, 400\rrbracket$) on a single rollout. Above, uniform coloring indicates adherence to fluid incompressibility. Qualitatively, \ours is able to remain close to the ground truth for the entire rollout.}
  % \label{fig:dam}
\end{figure}

As established in Fig~\ref{fig:dam}, even in short time frames, both GNS and SEGNN struggle to maintain fluid incompressibility, and thus, failed to learn robust physical patterns. We see that GNS and SEGNN introduce density issues even as early as $t = 20$. The density issues on GNS are especially apparent in the images for $t= 80$ and $t = 140$, and still remain prominent on SEGNN. By the time the fluid has hit the wall at the other side and formed a wave, we see \ours still maintains a relatively close distribution to the ground truth, whereas both GNS and SEGNN continue to struggle. Whereas before the GNS and SEGNN still predicted a similar silhouette, we see that at around $t = 200$ the shapes themselves become distorted. This is not a problem for \ours even up to $t = 380$. However, after $t = 200$, our model's density predictions do become slightly inconsistent with the ground truth, and we hypothesize that SPH relaxation may be used here to address it.

Our results on the other dataset in Table~\ref{tab:abl-long} show that in less turbulent cases, the performance of \ours is still comparable to NeuralSPH on long rollouts.

\begin{table}[H]
\centering
\caption{Performance for MSE400 as defined by \citet{toshev2024neural}. As we were unable to reproduce the results of NeuralSPH, we utilize their results for GNS and NeuralSPH.}
\label{tab:abl-long}
  \begin{tabular}{ccccccc}
  % --------------------------------

    \toprule
    \textbf{Dataset} & \textbf{Model} &
      \multicolumn{1}{c}{$\text{MSE}_{400}$} &
      \multicolumn{1}{c}{$\text{MSE}_{E_{\text{kin}}}$} &
      \multicolumn{1}{c}{Sinkhorn} \\
\midrule
\multirow{3}{*}{DAM-2D}
& GNS & $\num{1.9e-1}$ & $\num{4.6e-2}$ & $\num{3.8e-2}$ \\
& NeuralSPH &  $\num{8.4e-2}$ & $\num{2.1e-3}$ & $\num{7.5e-3}$ \\
& \ours & $\num{2.84e-2}$ & $\num{3.04e-4}$ & $\num{9.59e-4}$ \\
\midrule
\multirow{3}{*}{LDC-2D}
& GNS & $\num{3.3e-2}$ & $\num{1.1e-4}$ & $\num{3.1e-4}$ \\
& NeuralSPH &  $\num{1.6e-2}$ & $\num{1.2e-6}$ & $\num{2.8e-7}$ \\
& \ours & $\num{1.64e-2}$ & $\num{1.44e-4}$ & $\num{5.00e-5}$ \\
\midrule
\multirow{3}{*}{RPF-2D}
& GNS & $\num{2.7e-2}$ & $\num{4.3e-3}$ & $\num{3.7e-7}$ \\
& NeuralSPH &  $\num{2.7e-2}$ & $\num{1.4e-4}$ & $\num{3.0e-8}$ \\
& \ours & $\num{2.28e-2}$ & $\num{2.60e-4}$ & $\num{3.93e-7}$ \\
\midrule
\multirow{3}{*}{TGV-2D}
& GNS & $\num{5.3e-4}$ & $\num{5.6e-7}$ & $\num{5.4e-7}$ \\
& NeuralSPH &  $\num{4.8e-4}$ & $\num{4.8e-7}$ & $\num{1.7e-8}$ \\
& \ours & $\num{4.55e-4}$ & $\num{4.22e-7}$ & $\num{4.78e-7}$ \\
\midrule
\multirow{3}{*}{LDC-3D}
& GNS & $\num{3.2e-2}$ & $\num{1.3e-7}$ & $\num{2.0e-5}$ \\
& NeuralSPH &  $\num{3.2e-2}$ & $\num{2.9e-8}$ & $\num{1.1e-6}$ \\
& \ours & $\num{3.23e-2}$ & $\num{3.12e-8}$ & $\num{1.04e-6}$ \\
    \bottomrule
  \end{tabular}
\end{table}

\subsection{UPT}
As the UPT authors provide configurations to obtain MSE20 over the LagrangeBench datasets, we provide a comparison in Table~\ref{tab:abl-upt}.

\begin{table}[H]
\centering
\caption{Performance for MSE20 against UPT.}
\label{tab:abl-upt}
  \begin{tabular}{ccccccc}
  % --------------------------------

    \toprule
    \textbf{Model} & DAM-2D &
      LDC-2D &
      RPF-2D &
      TGV-2D &
      RPF-3D &
      TGV-3D \\
\midrule
UPT & $\num{4.28e-4}$ & $\num{3.34e-3}$ & $\num{1.12e-2}$ & $\num{1.11e-2}$ & $\num{5.33e-4}$ & $\num{1.06e0}$ \\
\ours & $\num{1.55e-5}$ & $\num{1.45e-5}$ & $\num{1.54e-6}$ & $\num{3.81e-6}$ & $\num{1.95e-5}$ & $\num{6.10e-3}$ \\
    \bottomrule
  \end{tabular}
\end{table}

We believe that this difference is due to the different objectives of the models. UPT is designed to model the velocity field directly, and hence is significantly weaker than \ours at predicting particle trajectories across all datasets. Hence, we do not find the models to be directly comparable, although \ours may be used for UPT's applications and UPT may be used for \ours's applications. In terms of inference time, the two methods are comparable: e.g., \ours averages 0.013 seconds per rollout step on TGV-2D, while UPT averages 0.014 seconds.

\section{Limitations and Future Work} \label{sec:lim}
\textbf{Resolution Coupling.} Our method's fixed Eulerian grid can oversmooth dense particle regions and inefficiently allocate resolution to sparse areas, acting as a low-pass filter. This uniform spacing struggles with non-uniform particle distributions, potentially losing sub-grid details. Adaptive grids (e.g., octrees, deformable kernels) could help, but maintaining the simplicity and efficiency of regular convolutions is an open challenge.

\textbf{Grid-Induced Anisotropy.} The axis-aligned grid introduces orientation bias, potentially representing features aligned with grid axes more sharply than others (e.g., diagonal shocks may ``stair-step''). Potential solutions include running multiple CNN branches with rotated projections or using group-equivariant/steerable convolutions for built-in rotational equivariance, with the trade-off being increased computational cost.

% \section*{LLM Usage}
% LLMs were used to polish writing and to help debug code.

\end{document}